\documentclass[11pt,fullpage]{article}
\usepackage{amsmath,graphicx,amsfonts,amssymb,sansseriftitles,bm} 
\usepackage[title]{appendix}
\usepackage{algorithm,algorithmic}
\usepackage{caption}
\usepackage[section]{placeins}
\usepackage{lscape}
\usepackage{xcolor}
\usepackage[colorlinks=true,citecolor=red,linkcolor=red,urlcolor=red]{hyperref}
\usepackage{subfigure}
\usepackage{setspace}
\usepackage{ragged2e}
\usepackage{multirow,booktabs}
\usepackage{thm-restate}

\usepackage{eufrak}
\usepackage{breqn}
\usepackage{mathtools}
\usepackage{bbm}

\usepackage[numbers,sort&compress]{natbib}

\usepackage{tikz}
\usetikzlibrary{patterns}
\usetikzlibrary{automata, positioning}
\usetikzlibrary{shapes,decorations,arrows,calc,arrows.meta,fit,positioning}
\tikzset{
    -Latex,auto,node distance =1 cm and 1 cm,semithick,
    state/.style ={circle, draw, minimum width = 0.8 cm},
    state2/.style ={circle, draw, minimum width = 0.1 cm, inner sep=0pt},
    point/.style = {circle, draw, inner sep=0.04cm,fill,node contents={}},
    bidirected/.style={Latex-Latex,dashed},
    el/.style = {inner sep=2pt, align=left, sloped}
}

\input epsf
\allowdisplaybreaks

\def\blue#1{{\color{blue}#1}}

\newcommand\leaveline{\vspace{0.3cm}}

\def\bi{\begin{itemize}}
\def\ei{\end{itemize}}
\def\bnum{\begin{enumerate}}
\def\enum{\end{enumerate}}

\newcommand{\bR}{\mathbb{R}}
\newcommand{\bI}{\mathbb{I}}

\newcommand{\bO}{\mathbb{O}}
\newcommand{\bH}{\mathbb{H}}

\newcommand{\bP}{\mathbb{P}}
\newcommand{\bE}{\mathbb{E}}

\newcommand{\cF}{\mathcal{F}}

\newcommand{\CP}{C^{\mbox{\scriptsize P}}}
\newcommand{\CN}{C^{\mbox{\scriptsize N}}}
\newcommand{\CI}{C^{\mbox{\scriptsize I}}}

\newcommand{\ThetaP}{\Theta^{\mbox{\scriptsize P}}}
\newcommand{\ThetaN}{\Theta^{\mbox{\scriptsize N}}}

\newcommand{\EWACP}{\text{EWAC}^{\mbox{\scriptsize P}}}
\newcommand{\EWACN}{\text{EWAC}^{\mbox{\scriptsize N}}}
\newcommand{\EWACI}{\text{EWAC}^{\mbox{\scriptsize I}}}
\newcommand{\WACP}{\text{WAC}^{\mbox{\scriptsize P}}}

\newcommand{\thetaP}{\theta^{\mbox{\scriptsize P}}}
\newcommand{\thetaN}{\theta^{\mbox{\scriptsize N}}}
\newcommand{\thetaI}{\theta^{\mbox{\scriptsize I}}}

\newcommand{\Pb}{\mathbb{P}}
\newcommand{\fQ}{\mathbf{Q}}
\newcommand{\fE}{\mathbf{E}}

\newcommand{\fU}{\mathbf{U}}
\newcommand{\fM}{\mathbf{M}}
\newcommand{\fV}{\mathbf{V}}

\newcommand{\fp}{\mathbf{p}}

\newcommand{\ftheta}{\pmb{\theta}}

\newcommand{\tO}{\widetilde{O}}
\newcommand{\tH}{\widetilde{H}}

\newcommand{\tildeo}{\widetilde{o}}

\newcommand{\tx}{\widetilde{x}}

\newcommand{\tY}{\widetilde{Y}}

\newcommand{\bpi}{\bm{\pi}}
\newcommand{\btheta}{\bm{\theta}}
\newcommand{\bPi}{\bm{\Pi}}

\newcommand{\bU}{\bm{U}}

\newcommand{\wobs}{w_{\text{\footnotesize obs}}}

\newcommand{\Tmax}{T_{\mbox{\tiny max}}}
\newcommand{\fair}{\text{f}}
\newcommand{\loaded}{\text{b}}

\newtheorem{defi}{Definition}
\newtheorem{thm}{Theorem}

\newtheorem{rem}{Remark}

\newcommand*{\Scale}[2][4]{\scalebox{#1}{$#2$}}%
\definecolor{mhred}{RGB}{200,21,21}

\graphicspath{ {images/} }

\begin{document}

\title{\vspace{-5mm}\sffamily{\bfseries A Counterfactual Analysis of the \\ Dishonest Casino} \vspace{0.3cm}}

\author{\sffamily \textbf{Martin B. Haugh\footnote{Corresponding author.}} \\
\sffamily  Department of Analytics, Marketing \& Operations \\
\sffamily  Imperial College Business School, Imperial College \\
\texttt{m.haugh@imperial.ac.uk}
\vspace{.3cm}
\and \sffamily \textbf{Raghav Singal} \\
\sffamily  Operations and Management Science \\
\sffamily   Tuck School of Business, Dartmouth College \\
\texttt{singal@dartmouth.edu}}

\date{\today}

\maketitle

\centerline {\large \textbf{Abstract}} 
\noindent
The dishonest casino is a well-known hidden Markov model (HMM) often used in education to introduce HMMs and graphical models. A sequence of die rolls is observed with the casino switching between a fair and a loaded die. Instead of recovering the latent regime through filtering, smoothing, or the Viterbi algorithm, we ask a counterfactual question: how much of the gambler’s winnings are caused by the casino's cheating? We introduce a class of structural causal models (SCMs) consistent with the HMM and define the expected winnings attributable to cheating (EWAC). Because EWAC is only partially identifiable, we bound it via linear programs (LPs). Numerical experiments help to develop intuition using benchmark SCMs based on independence, comonotonic, and countermonotonic copulas. Imposing a time homogeneity condition on the SCM yields tighter bounds, whereas relaxing it produces looser bounds that admit an explicit LP solution. Domain knowledge such as pathwise monotonicity or counterfactual stability can be incorporated through additional linear constraints. Finally, we show the time averaged EWAC becomes fully identifiable as the number of time periods tends to infinity. Our work is the first to develop LP bounds for counterfactuals in a HMM setting, benefiting educational contexts where counterfactual inference is taught.



\clearpage

\section{Introduction} \label{sec:intro}
The so-called {\em dishonest casino} is a well-known example of a hidden Markov model (HMM) that is used in many educational settings as an introduction to HMMs and graphical models more generally. In this setting, a finite sequence of die rolls is observed but unbeknownst to observers, the casino periodically switches between using a fair die and a loaded die.  The typical goal is to then use the observed sequence of die rolls to infer the pattern of fair / loaded dice that generated the die rolls. The example is widely used in teaching HMMs because it effectively demonstrates the concepts of states, observations, transitions, and emissions which then lead  naturally to filtering and smoothing algorithms as well as the Viterbi algorithm for finding the most likely path of the hidden states. Indeed a Google search of ``dishonest casino'' AND ``HMM'' yields over 2,000 hits including many links to university course web pages that teach HMMs. It is also used in popular textbooks, e.g., \cite{pml2Book,DEKM1998,WoolfEtAl}, and as a motivating example in documentation for popular HMM libraries in Python and R, e.g., the \texttt{hmmlearn} and \texttt{ghmm} Python libraries and the \texttt{hmm} R library.

In this paper, we go beyond the standard inference questions regarding the dishonest casino and consider how much of the casino's winnings is attributable to cheating given an observed sequence of die rolls. This is inherently a counterfactual question and cannot be answered using the primitives of the HMM, i.e., the initial, transition, and emission distributions. In order to answer this question, we need a structural causal model (SCM) which explicitly models the data-generating mechanism of the dishonest casino HMM. We introduce a natural class of SCMs that are consistent with the HMM and show that the expected winnings attributable to cheating (EWAC) within that class can be bounded via linear programs (LPs). We typically impose a natural  time-homogeneity condition on the SCM and in a series of numerical experiments, we compute the LP bounds and develop intuition for them via some benchmark SCMs that are based on the independence, comonotonic, and counter-monotonic copulas. When we don't impose time-homogeneity on the SCM, we obtain looser bounds but show that the LPs decouple with respect to time and allow for an almost explicit solution of the LPs. We also show that domain specific knowledge such as pathwise monotonicity or the recently proposed counterfactual stability property can easily be imposed via additional linear constraints in the LPs.  All of the bounds that we compute are the tightest possible for the given assumptions, e.g., time homogeneity, counterfactual stability, etc. While the EWAC is only partially identifiable (and can therefore only be bounded) when the number of die rolls $T$ is finite, we use results from the theory of Markov chains to also show the time-average EWAC is fully identifiable in the limit as the number of time periods goes to infinity.

Our work contributes to the literature on bounding counterfactuals in causal inference, and to the best of our knowledge, we are the first to develop LP bounds in a dynamic setting such as an HMM. Because of the stylized nature of the dishonest casino and its ubiquity in educational settings, this work might be of particular benefit in such settings where counterfactual inference is taught.

While the dishonest casino is admittedly stylized, it provides a canonical and deliberately minimal dynamic latent-state model in which the distinction between a probabilistic description and a counterfactual attribution query is especially transparent. In many domains, practitioners use an HMM (or a close variant) to represent an unobserved regime $H_t$ generating observable signals $O_t$, together with an additive performance metric or reward $w_{O_t}$. Even when the HMM primitives are known or well-estimated, retrospective questions of the form ``how much of the realized cumulative reward is attributable to a particular mechanism/intervention?'' are not answered by the HMM alone: they require an SCM that couples potential outcomes across regimes and thereby pins down (or partially identifies) the relevant counterfactual world. Our dishonest-casino analysis should therefore be viewed as a worked, fully sharp example of partial identification in a dynamic latent-state setting, where the resulting bounds admit clean intuition and where additional domain restrictions (e.g., monotonicity or counterfactual stability) can be incorporated transparently. 

More broadly, the dishonest-casino model is best understood as a tractable base case of a general counterfactual program for dynamic latent-state models with richer hidden-state structure and with policy- or action-level interventions. For example, generalized HMMs (GHMM) with exogenous inputs allow emissions to depend on $(H_t,X_t)$ and state transitions to depend on $(H_t,O_t)$, which is a natural template in sequential decision problems (e.g., dynamic treatment regimes and POMDP-style models), where the intervention changes an entire policy sequence $X_{1:T}$. In related work, \cite{HaughSingal2024} adopts such a GHMM framework to study policy interventions in a healthcare setting. Specifically, they consider a setting where a woman had been incorrectly denied periodic breast cancer screenings for a period of time during which she developed and died from breast cancer. They bound a probability of necessity, i.e., the counterfactual probability the woman would have survived had the periodic screenings been permitted. Their work requires combining hidden-state inference with polynomial optimization over a natural space of SCMs consistent with the observed GHMM. The dishonest-casino setting of this paper is a much simpler one that isolates the core identification and modeling issues in their simplest form (and yields LP-based sharp bounds).

Our work is directly related to the work on partial identification in causal inference.
In early work, \cite{Manski1990} produces analytic bounds on counterfactual quantities which in practice are often loose. The seminal work of \cite{BALKE-Pearl94} showed that sharp bounds can be computed as the solutions of LPs and was later extended by \cite{TianPearl2000}. Unfortunately, many partial identification problems do not fit in their framework. Analytic non-linear solutions for specific types of problems have also been obtained, e.g., \cite{Kennedy2019,knox2020administrative,Gabriel2022}.  More recently, \cite{duarte2023automated} extend results from \cite{Geiger_UAI1999} and \cite{Wolfe2019} to argue that essentially all discrete partial identification problems can be formulated as polynomial optimization problems whose solutions will provide the tightest possible bounds. They solve polynomial programs and succeed in recovering (and sometimes improving on) previously known bounds for specific problem types. In other recent work,  \cite{Zhang-Tian-Barenboim-2021} develop an approach that takes an arbitrary causal model with discrete observed variables as input, and then returns a canonical model that is equivalent in terms of both observational and counterfactual distributions. They also recognize that tight counterfactual bounds can be obtained via polynomial programs. Unlike \cite{duarte2023automated}, however, they do not solve the polynomial programs and instead, they propose MCMC algorithms to approximate their solutions.

The remainder of this paper is organized as follows. In Section \ref{sec:CasinoOpt}, we introduce the dishonest casino HMM as well as the counterfactual query regarding the winnings of the casino. We introduce a natural class of SCMs that are consistent with the HMM in Section \ref{sec:CasinoSCM} and then derive our linear-programming based bounds for the EWAC in Section \ref{sec:CasinoLPs}. We also show in Section \ref{sec:CasinoLPs} that the time-average EWAC is fully identifiable in the limit as the number of time periods goes to infinity. In Section \ref{sec:PM_CS}, we discuss how domain-specific knowledge in the form of pathwise monotonicity and counterfactual stability can be modeled via linear constraints, while in Section \ref{sec:CopulaEGS} we introduce some simple benchmark models that are based on the well known independence, comonotonic and countermonotonic copulas.
Section \ref{sec:CasinoExp} describes our numerical experiments and we conclude in Section \ref{sec:conc}.  Appendix \ref{sec:CopulaAppendix} provides a very brief introduction to copulas while Appendix \ref{sec:AdditionalNums} contains some additional numerical results.

\section{The Dishonest Casino and the Counterfactual Query}  \label{sec:CasinoOpt}
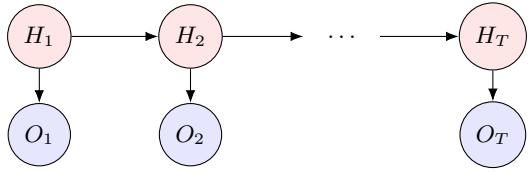
\begin{figure}[ht]
\centering
\begin{tikzpicture}[scale=1.0,transform shape]
    \node[state,fill=red!10!white] (h1) at (0,0) {\footnotesize $H_1$};
    \node[state,fill=red!10!white] (h2) at (2,0) {\footnotesize $H_2$};
    \node[draw=none, minimum width = 1 cm] (hdots) at (4,0) {\footnotesize $\ldots$};
    \node[state,fill=red!10!white] (hT) at (6,0) {\footnotesize $H_T$};
    \node[state,fill=blue!10!white] (o1) at (0,-1.3) {\footnotesize $O_1$};
    \node[state,fill=blue!10!white] (o2) at (2,-1.3) {\footnotesize $O_2$};
    \node[state,fill=blue!10!white] (oT) at (6,-1.3) {\footnotesize $O_T$};
    \path (h1) edge (o1);
    \path (h1) edge (h2);
    \path (h2) edge (o2);
    \path (h2) edge (hdots);
    \path (hT) edge (oT);
    \path (hdots) edge (hT);
\end{tikzpicture}
\caption{The dishonest casino. The states $H_{1:T}$ represent the fair / biased die and are hidden while the emissions $O_{1:T}$,  i.e., die rolls, are observed.}
\label{fig:SCM-Casino1}
\end{figure}

In this section, we describe the dishonest casino HMM and then consider the counterfactual query that we wish to answer. The dishonest casino is a $T$-period HMM with hidden states and observations denoted by $H_t \in \bH$ and $O_t \in \bO$ for time $t \in [T]$, respectively.  The hidden state $H_t \in \{\fair, \loaded\}$ for $t \in [T]$ denotes whether the casino is using a {\em fair} die ($H_t=\fair$) or a {\em biased} / \emph{loaded} die ($H_t=\loaded$).  Under a fair die, each of the six observations or die rolls ($O_t \in \{1,\ldots,6\}$) is equally likely whereas under a loaded die,  higher observations are more likely.   The standard graphical model for HMMs is displayed in Figure \ref{fig:SCM-Casino1} and it encapsulates the various dependence / independence relationships in the HMM. For example, the $H_t$'s form a Markov chain, i.e.,  $H_t \mid H_{1:(t-1)} = H_t \mid H_{t-1}$, and $O_t \mid H_t$ is independent of all other hidden states and observations. (We use $X\mid Y$ throughout to denote the distribution of $X$ given $Y$.) The casino earns a reward of $w_i \in \bR$ if the time $t$ die roll is $O_t = i$. W.l.o.g., we shall assume that $w_i$ is increasing in $i$ reflecting the fact that the casino expects to win more when using the loaded die.

The model $\fM \equiv (\fp,  \fE,  \fQ)$ comprises three primitives, namely the initial state distribution $\fp := [p_h]_h$ where $p_{h} := \Pb(H_1 = h)$, the emission probability matrix $\fE := [e_{hi}]_{h \in \{1,2\}, i \in \{1,\ldots,6\}}$ where $e_{hi} := \bP(O_{t} = i \mid H_t = h)$, and the hidden-state transition probability matrix $\fQ := [q_{hh'}]_{h,h' \in \{1,2\}}$ where $q_{hh'} := \bP(H_{t+1} = h' \mid H_t = h)$.
We occasionally use $h$ as an index in vectors and matrices in which case, we use $h=1$ to denote the fair state and $h=2$ to denote the biased state.
Each of these quantities is assumed to be known.  (Given sufficient data, one can estimate these quantities using standard HMM estimation techniques such as the Baum-Welch algorithm.)

Given an observation sequence $O_{1:T}:=\{O_1, \ldots , O_T \}$, we can use standard filtering and smoothing algorithms to compute the posterior distribution of $H_t \mid O_{1:T}$ for any $t$. It is also straightforward to simulate posterior sample paths of $H_{1:T} \mid O_{1:T}$ or to compute the most likely hidden path given $O_{1:T}$. We refer the reader to \cite{barber2012bayesian} for a description of these algorithms as well as further details on HMMs and their extensions.

\subsubsection*{The Counterfactual Query}
Suppose we observe a sequence of die rolls $o_{1:T}$  implying that the casino won
\begin{equation} \label{eq:Wobs}
\wobs := \sum_{t=1}^T w_{o_t}.
\end{equation}
We can imagine the cheating of the casino has been discovered and a court now wishes to compute its {\em expected winnings attributable to cheating}, or EWAC. That is, the court wants to determine the difference between what the casino won and how much it would expect to have won had it been forced to use the fair die at all times {\em given} the sequence of observed winnings while employing the Markov policy of switching back and forth between the fair and loaded dice.  The key aspect of (\ref{eq:Wobs}) that enables us to bound the EWAC via linear programs is that $\wobs$ is additive across time. We could also include a constant term $C$ so that $\wobs := C+\sum_{t=1}^T w_{o_t}$. In that case, $C$ would be present in both $\wobs$ and the expected counterfactual winnings, and would therefore have no impact on EWAC as we can see from (\ref{eq:EWACDefn}) below.

It's important to note this question cannot be answered by simply calculating the expected winnings of the casino in a fresh run of $T$ die rolls where it is forced to use the fair die at all times. Indeed the casino's expected winnings in this scenario is simply $T \times  (\sum_{i=1}^6w_i)/6$. But simply taking $\wobs - T \times (\sum_{i=1}^6w_i)/6$ as our EWAC is incorrect as it does not condition on the observed sequence of die rolls $o_{1:T}$ to infer (among other things) just how often the casino actually used the loaded die. For example, if the observed die rolls were a sequence of 1's, then that would imply (given our earlier assumption of the loaded die favoring higher die rolls) that the casino probably used the fair die considerably more often than might otherwise have been expected given the HMM dynamics.
In this case, the EWAC might be relatively small or even negative! Indeed, in the extreme case where throwing a 1 under the loaded die is impossible,  we could conclude that the casino, however unlikely, always used the fair die for the observed sequence and so the EWAC was zero. Similarly, an observed sequence with a relatively large number of 6's might suggest the casino used the loaded die far more frequently than might otherwise have been expected and hence, the EWAC might have been large. It is therefore necessary to account for the role of {\em luck} as evidenced by $o_{1:T}$ in answering the counterfactual query. This will be the key ``abduction'' step in our analysis.

Before defining EWAC mathematically, we must introduce one further piece of notation. We will use ${\cal P}_{\text{dis}}$ to denote the policy of occasionally using the loaded die according to the HMM defined by $\fM \equiv (\fp,  \fE,  \fQ)$. Similarly, we will use ${\cal P}_{\text{fair}}$ to denote the policy of {\em always} using the fair die. We can now define the EWAC as
\begin{subequations}
\label{eq:EWACDef}
\begin{align}
\text{EWAC} := \wobs - \sum_t \bE [w_{\tO_t}]
\label{eq:EWACDefn}
\end{align}
where the \emph{counterfactual} observations are defined as
\begin{align}
\tO_{1:T} := O_{1:T}^{{\cal P}_{\text{fair}}}  \mid (O_{1:T} = o_{1:T}, {\cal P}_{\text{dis}}).
\label{eq:tilde}
\end{align}
\end{subequations}
That is, $\tO_{1:T}$ is the sequence of die rolls that would have been observed under the policy ${\cal P}_{\text{fair}}$, given that the actual policy ${\cal P}_{\text{dis}}$ was in place when $O_{1:T} = o_{1:T}$ was observed.
Throughout the paper, we will use ``tilde'' notation to concisely represent quantities in the counterfactual world. 
It's clear from (\ref{eq:EWACDefn}) that we only need the marginal distribution of the $\tO_{t}$'s in order to compute the EWAC for a given SCM (as defined in Section \ref{sec:CasinoSCM} below).  To do this, we need to condition on the observed die rolls $o_{1:T}$ and ${\cal P}_{\text{dis}}$ (the abduction step), and then set the policy to ${\cal P}_{\text{fair}}$ in the counterfactual world (the action step).
Finally, the prediction step will then require us to estimate the winnings of the casino in the counterfactual world. Together,  abduction, action, and prediction are the steps required to answer a counterfactual query; see, for example, \cite{pearl_2009,pearl_2018}. We will execute all three steps in Section \ref{sec:CasinoLPs} but first we must define a structural causal model for the HMM.

\section{The Structural Causal Model}  \label{sec:CasinoSCM}

\begin{figure}[ht]
\centering
\begin{tikzpicture}[scale=1.0,transform shape]
    \node[state,fill=red!10!white] (h1) at (0,0) {\footnotesize $H_1$};
    \node[state,fill=red!10!white] (h2) at (2,0) {\footnotesize $H_2$};
    \node[state2,fill=black!10!white] (u1) at (-1,0.7) {\footnotesize $U_1$};
    \node[state2,fill=black!10!white] (u2) at (1,0.7) {\footnotesize $\fU_2$};
    \node[draw=none, minimum width = 1 cm] (hdots) at (4,0) {\footnotesize $\ldots$};
    \node[state,fill=red!10!white] (hT) at (6,0) {\footnotesize $H_T$};
    \node[state2,fill=black!10!white] (uT) at (5,0.7) {\footnotesize $\fU_T$};
    \node[state,fill=blue!10!white] (o1) at (0,-1.3) {\footnotesize $O_1$};
    \node[state2,fill=black!10!white] (v1) at (-1,-1.8) {\footnotesize $\fV_1$};
    \node[state,fill=blue!10!white] (o2) at (2,-1.3) {\footnotesize $O_2$};
    \node[state2,fill=black!10!white] (v2) at (1,-1.8) {\footnotesize $\fV_2$};
    \node[state,fill=blue!10!white] (oT) at (6,-1.3) {\footnotesize $O_T$};
    \node[state2,fill=black!10!white] (vT) at (5,-1.8) {\footnotesize $\fV_T$};
    \path (u1) edge (h1);
    \path (h1) edge (o1);
    \path (h1) edge (h2);
    \path (h2) edge (o2);
    \path (h2) edge (hdots);
    \path (hT) edge (oT);
    \path (hdots) edge (hT);
    \path (u2) edge (h2);
    \path (uT) edge (hT);
    \path (v1) edge (o1);
    \path (v2) edge (o2);
    \path (vT) edge (oT);
\end{tikzpicture}
\caption{The SCM underlying the dishonest casino's HMM.  The only difference between the SCM here and the HMM of Figure \ref{fig:SCM-Casino1} is the addition of (grey) exogenous noise nodes $[\fU_t, \fV_t]_t$. }
\label{fig:SCM-Casino2}
\end{figure}
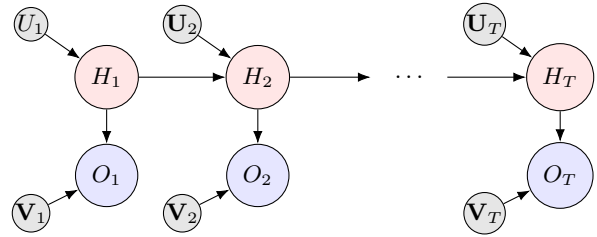

In order to answer any counterfactual query, we need to embed the HMM in an SCM; see, e.g., \cite{pearl_2009}. A representative SCM for the dishonest casino setting is shown in Figure \ref{fig:SCM-Casino2}.
Consider for example, the observation $O_t$ for any time $t$.  $O_t$ is a stochastic function of its parent $H_t$ and we may assume the stochasticity is driven by the exogenous noise vector $\fV_t := [V_{th}]_{h}$ which consists of $\lvert\bH\rvert $ noise random variables.
We model the exogenous noise as a vector (as opposed to a scalar) to capture the fact that each $O_{th} := O_t \mid (H_t = h)$ defines a \emph{distinct} random variable for all $h$. These random variables might be independent, or they might display positive or negative dependence.
One way to handle these various possibilities is to associate each $O_{th}$ with a distinct noise variable $V_{th}$.  The {\em dependence structure} among these noise variables $[V_{th}]_{h}$ is then what determines the dependence structure among $[O_{th}]_{h}$.  The structural equation obeys
\begin{subequations}
\label{eq:SCM_Casino}
\begin{align}
\Scale[1]{O_t = f(H_t, \fV_t) = \sum_{h} f_{h}(V_{th}) \bI\{H_t=h\}},
\label{eq:SCM_Casino_one}
\end{align}
where $f_{h}(\cdot)$ is defined using the emission distribution $[e_{hi}]_i$ and we assume wlog that $V_{th} \sim \text{U}[0,1]$.
Similarly,  for $t>1$, recognizing that each $H_{th} := H_{t} \mid (H_{t-1} = h)$ is a distinct random variable for all $h$, we associate each $H_{th}$ with its own noise variable $U_{th}$:
\begin{align}
\Scale[1]{H_t = g(H_{t-1}, \fU_t) = \sum_{h} g_{h}(U_{th}) \bI\{H_{t-1}=h\}},
\label{eq:SCM_Casino_two}
\end{align}
\end{subequations}
where $g_{h}(\cdot)$ is defined using the transition distribution $[q_{hh'}]_{h'}$ and $U_{th} \sim \text{U}[0,1]$ wlog. $\fU_t := [U_{th}]_{h}$ therefore consists of $\lvert\bH\rvert $ $\text{U}[0,1]$ random variables.  (Because there is no $H_0$, we just need a single uniform random variable $U_1$ to generate $H_1$. This is reflected in Figure \ref{fig:SCM-Casino2} where we do not use bold font for $U_1$.)

The representation in \eqref{eq:SCM_Casino_one} allows us to model $[O_{th}]_{h}$ and capture any dependence structure among these random variables by specifying the joint multivariate distribution of $\fV_t$.  Since the univariate marginals of $\fV_t$ ($\fU_t$) are known to be $\text{U}[0,1]$, specifying the multivariate distribution of $\fV_t$ ($\fU_t$) amounts to specifying the dependence structure or {\em copula} of $\fV_t$ ($\fU_t$).  For example, if the $V_{th}$'s are mutually independent (the independence copula) and we have $H_t=h'$,  then inferring the conditional distribution of $V_{th'}$ will tell us nothing about the $V_{th}$'s for $h \neq h'$.  Alternatively, if $V_{th} = V_{th'}$ for all $h$ and $h'$, then this models perfect positive dependence (the comonotonic copula) and inferring the conditional distribution of $V_{th'}$ amounts to simultaneously inferring the conditional distribution of all the $V_{th}$'s.

A careful reader might note that we could model the joint distributions of each of $[O_{th}]_{h}$ and $[H_{th}]_{h}$ {\em directly} rather than introducing the $\fV_t$'s and $\fU_t$'s and then using (\ref{eq:SCM_Casino_one}) and (\ref{eq:SCM_Casino_two}). This is indeed the case and in fact is well-known in the causal modeling community. Moreover, we shall take this approach in Section \ref{sec:CasinoLPs} below when we construct our LP bounds. However, for the remainder of this section, we will persist with our representation of the SCM in terms of the $\fV_t$'s and $\fU_t$'s as we find it helpful in aiding our understanding of SCM modeling. Towards this end, we make three additional observations.

\begin{enumerate}
\item An SCM is a {\em generative} model. In particular, because each component of $\fU_t$ and $\fV_t$ is $\text{U}[0,1]$, we could use them via the  inverse-transform approach from Monte-Carlo simulation  \citep[e.g.][]{GlynnAsmussen2007}
    to generate the relevant random variables. For example, the  $f_h(\cdot)$ function in (\ref{eq:SCM_Casino_one}) is in fact the inverse CDF function we use to generate $O_t$ given $H_t=h$ using $V_{th} \sim \text{U}[0,1]$. (It is for this reason the $\text{U}[0,1]$ assumption stated above is wlog.)

\item We emphasize the need to work with the exogenous vectors $(\fU_t,\fV_t)$ (or equivalently, the joint distributions of $[O_{th}]_{h}$ and $[H_{th}]_{h}$) when doing a {\em counterfactual} analysis since different joint distributions of $(\fU_t,\fV_t)$ will lead to (possibly very) different values of EWAC. If we were not doing a counterfactual analysis and only cared about the joint distribution of (a subset of) $(O_{1:T}, H_{1:T})$, then our analysis would only depend on the joint distribution of the $(\fU_t,\fV_t)$'s via their known univariate marginals.  In this case, it would not be necessary to describe or even mention the class of SCMs consistent with the HMM.

\item \label{comment:SimuateDieComon}
Continuing on from the previous point, if we just wanted to simulate the HMM (and not answer a counterfactual question), then specifying the SCM would be an overkill in that there would be no need to use a different random variable $V_{th}$ for each possible hidden state value $h$ when generating $O_t$. Indeed, we could just use a single random variable $V_{t}$ to generate $O_t \mid H_t$ regardless of the value of $H_t$. This would amount to implicitly assuming the comonotonic copula for the $V_{th}$'s as described above and it would be perfectly natural to do so as one and only one value of $H_t$ can ever occur on any simulated path. However, when answering a counterfactual query, we need to be able to imagine more than one value of $H_t$ occurring {\em simultaneously} on the same path: namely the (unknown) value of $H_t$ that occurred on the {\em observed} path and the {\em counterfactual} (again unknown) value $\tH_t$ that would occur in the counterfactual world on the same path. (It helps in this context to consider a ``path'' to be a single realization of $(U_1,\fU_{2:T},\fV_{1:T})$. The discussion here applies equally to the $U_{th}$'s used to generate $H_t \mid (H_{t-1}=h)$.)
\end{enumerate}

\noindent
Finally, we assume the collection of exogenous noise vectors $\{\fU_t,\fV_t\}_{t=1}^T$ is independent across time,
and the noise driving state transitions ($\fU_t$) is independent of the noise driving emissions
($\fV_{t'}$) for all $(t,t')$. This is a very natural assumption and is sufficient to recover the conditional independence relationships among the endogenous variables $(H_{1:T},O_{1:T})$ implied by the HMM of Figure \ref{fig:SCM-Casino1}.
Referring back to our Monte-Carlo simulation of the HMM analogy, this makes intuitive sense. For example, the $\text{U}[0,1]$ random variable $V_{t}$ used to generate $O_t \mid H_t$ ought to be independent of all the other uniform random variables that are used to generate the other emissions and state transitions. Our independence assumption regarding the exogenous noise vectors is a key feature of the SCM that allows us to construct our LP bounds. \cite{HaughSingal2024} invoked a similar mutual independence argument in formulating their polynomial programs to bound the probability-of-necessity in their modeling of the development and screening of breast cancer. 
We also note, however, that while this assumption defines a convenient and natural subclass of SCMs compatible with the HMM, it is not a necessary subclass. In particular, relaxing it can preserve the same observational HMM distribution while generally changing counterfactual quantities. For example, one could construct observationally equivalent SCMs by introducing shared exogenous randomness across components that are never jointly active under the realized parent values, e.g., by setting $V_{t \fair}=U_{(t+1) \loaded}$. But such constructions are considerably less compelling and we would not be able to construct our EWAC bounds were we to allow such SCMs. The class of SCMs we do consider is still very rich, however, since we allow dependence among the components of $\fV_t=[V_{th}]_h$ and among the components of $\fU_t=[U_{th}]_h$.

\section{Bounding EWAC via Linear Programs}  \label{sec:CasinoLPs}
We are now in a position to establish our LP bounds on the EWAC. We begin in Section \ref{sec:TH_LP} where we will impose a time-homogeneity constraint on the SCM. We will then relax this constraint in Section \ref{sec:TIH_LP}.  Finally,  in Section \ref{sec:TtoInf}, we establish that the time-average EWAC is completely identifiable as $T \to \infty$.

\subsection{The Time-Homogeneous Case} \label{sec:TH_LP}
While specifying the SCM in terms of the $\fU_t$'s and $\fV_t$'s is useful from a conceptual point of view, it's convenient to work with a more direct but equivalent construction of the SCM. Consider for example the relationship between $\fV_t := [V_{th}]_{h}$ and $[O_{th}]_{h}$. We know from (\ref{eq:SCM_Casino_one}) that the joint distribution of $[O_{th}]_{h}$ is completely determined by the joint distribution of $\fV_t$.
But we only care about the joint distribution of $\fV_t$ to the extent that it controls the joint distribution of $[O_{th}]_{h}$. It therefore makes more sense to model the joint distribution of $[O_{th}]_{h}$ {\em directly} rather than indirectly via the joint distribution of $\fV_t$.  (Indeed, there will be infinitely many joint distributions of $\fV_t$ that all lead to the same joint distribution of $[O_{th}]_{h}$. This is a consequence of Sklar's Theorem from the theory of copulas and is further discussed  in Appendix \ref{sec:CopulaAppendix}.)

Towards this end, let $\theta(i,j) := \Pb( O_{t\fair} = i,  O_{t\loaded} = j)$  be the bivariate PMF of $(O_{t\fair},  O_{t\loaded})$, where we recall that $O_{th} := O_t \mid (H_t = h)$. We assume that $\theta$ is independent of $t$, which amounts to a time-homogeneity constraint on the SCM. In our counterfactual world, the casino is forced to use the fair die at all times $t$ so that $\tH_{1:T} = \fair$. The hidden states in the counterfactual world are therefore deterministic and so the joint distribution of the exogenous noise vectors $\bU_t$ from Figure \ref{fig:SCM-Casino2} becomes irrelevant. This means there is no need to define a joint distribution over $[H_{th}]_{h}$.

We also define $\delta_t(h) := \Pb(H_t=h \mid o_{1:T}, {\cal P}_{\text{dis}})$, which is easily computed via standard filtering and smoothing algorithms for HMMs; see e.g., \cite{barber2012bayesian}. We are now ready to characterize the EWAC (recall \eqref{eq:EWACDefn}) for any time-homogeneous SCM. We do so in Proposition \ref{prop:EWACCharacterization} and we prove it here as the proof is informative and shows how information flows from the observed sequence of die rolls $o_{1:T}$ all the way through to the calculation of EWAC.

\begin{restatable}[\textbf{EWAC Characterization}]{prop}{EWACCharacterization} \label{prop:EWACCharacterization}
For any time-homogeneous SCM $\btheta :=\{\theta(i,j)\}_{i,j}$, we have
\begin{align}
\Scale[1]{\text{EWAC}(\btheta) = \wobs - \sum_{t=1}^T \left\{ \delta_t(\fair) \times w_{o_t} + \delta_t(\loaded) \times \sum_{i=1}^6 w_i \frac{\theta(i,o_t)}{e_{\loaded o_t}} \right\}.}
\label{eq:EWAC}
\end{align}
\end{restatable}

\paragraph{Proof} Recall from \eqref{eq:EWACDefn} that $\text{EWAC} := \wobs - \sum_{t=1}^T \bE [w_{\tO_t} ]$. We also have
\begin{subequations}
\begin{align}
\bE [w_{\tO_t} ] &= \bE \left[w_{O_{t}^{{\cal P}_{\text{fair}}}\mid o_{1:T}, {\cal P}_{\text{dis}} } \right] \label{eq:CFWinning10}  \\
&= \bE [w_{O_{t\fair}} \mid o_{1:T}, {\cal P}_{\text{dis}}] \label{eq:CFWinning0} \\
&= \bE [w_{O_{t\fair}} \mid o_{1:T}, {\cal P}_{\text{dis}}, H_t=\fair] \, \Pb(H_t=\fair \mid o_{1:T}, {\cal P}_{\text{dis}}) \nonumber \\
& \ \ \ \  + \bE [w_{O_{t\fair}} \mid o_{1:T}, {\cal P}_{\text{dis}}, H_t=\loaded]  \, \Pb(H_t=\loaded \mid o_{1:T}, {\cal P}_{\text{dis}}) \label{eq:CFWinning01}  \\
& = \bE [w_{O_{t\fair}} \mid o_{1:T}, H_t=\fair] \, \delta_t(\fair) + \bE [w_{O_{t\fair}}\mid o_{1:T},  H_t=\loaded] \, \delta_t(\loaded) \nonumber \\
&= w_{o_t}  \, \delta_t(\fair) + \underbrace{\bE [w_{O_{t\fair}} \mid o_{1:T},  H_t=\loaded]}_{=: (\star)}   \delta_t(\loaded), \label{eq:CFWinning}
\end{align}
\end{subequations}
where (\ref{eq:CFWinning0}) follows since $O_{t}^{{\cal P}_{\text{fair}}} = O_{t\fair}$, (\ref{eq:CFWinning01}) follows from the law of total probability and conditioning on $H_t$, and
(\ref{eq:CFWinning}) follows since $\bE [w_{O_{t\fair}} \mid o_{1:T}, H_t=\fair] =  w_{o_t} $ by the consistency axiom.
The only term in (\ref{eq:CFWinning}) that remains to compute is the term labelled $(\star)$. We obtain
\begin{subequations}
\begin{align}
(\star) := \bE [w_{O_{t\fair}} \mid o_{1:T},  H_t=\loaded]   &= \bE [w_{O_{t\fair}} \mid o_{t},  H_t=\loaded] \label{eq:Star000} \\
&= \bE [w_{O_{t\fair}} \mid O_{t\loaded} = o_t ] \nonumber \\
   &= \sum_{i=1}^6 w_i \times \Pb( O_{t\fair} = i \mid O_{t\loaded} = o_t ) \label{eq:Star00} \\
   &= \sum_{i=1}^6 w_i \times \frac{\Pb( O_{t\fair} = i, O_{t\loaded} = o_t )}{\Pb( O_{t\loaded} = o_t )} \nonumber \\
   &= \sum_{i=1}^6 w_i \times \frac{\theta(i,o_t)}{e_{\loaded o_t}},  \label{eq:Star}
\end{align}
\end{subequations}
where (\ref{eq:Star000}) follows since, given $H_t$, $O_{t\fair}$ only depends on $O_{1:T}$ through $O_t$. Combining \eqref{eq:EWACDefn}, \eqref{eq:CFWinning}, and \eqref{eq:Star}, we obtain \eqref{eq:EWAC}.
\hfill $\Box$

\leaveline

It's perhaps worth noting that the three steps required for any counterfactual analysis, namely the abduction, action and prediction steps (see, for example, \cite{pearl_2009}), are all executed within the proof of Proposition \ref{prop:EWACCharacterization}. The abduction step conditions on the observed sequence of die rolls to compute (i) the $\delta_t(h)$'s and (ii) the conditional probability $\Pb( O_{t\fair} = i \mid O_{t\loaded} = o_t)$ terms in (\ref{eq:Star00}). The action step imposes the policy ${\cal P}_{\text{fair}}$, i.e., it forces the fair die to be used at all times on the counterfactual path,  and is explicit in our definition of $w_{\tO_t}$ that we use in (\ref{eq:CFWinning10}). Finally, the prediction step then computes the expected value of the counterfactual, i.e., the EWAC.

The only unknown in (\ref{eq:EWAC}) is $\btheta$ but its one-dimensional marginals are known and given by the appropriate emission distribution.
One possibility is to specify $\btheta$ via a copula such as the independence, comonotonic, or countermonotonic copulas for example (see Section \ref{sec:CopulaEGS} where we discuss these benchmark SCMs), but selecting an appropriate copula is difficult even with detailed domain-specific knowledge. Instead, we will obtain upper and lower bounds on the EWAC by treating $\btheta$ as a matrix of decision variables. The feasible set
\begin{align*}
\Scale[1]{\mathcal{F} := \left\{\btheta \ge 0: \sum_{j} \theta(i,j) = e_{\fair  i} \ \forall i, \sum_{i} \theta(i,j) = e_{\loaded j} \ \forall j\right\}}
\end{align*}
ensures  $\btheta$ is a PMF whose marginals coincide with the known emission probabilities.  Since \eqref{eq:EWAC} is linear in $\btheta$ and so are all the constraints in $\mathcal{F}$, our upper and lower bounds on the EWAC are obtained as the solutions to the following linear programs:
\begin{subequations}
\begin{align*}
\text{EWAC}^{\text{ub}} &:= \max_{\btheta \in \mathcal{F}} \ \ \text{EWAC}(\btheta)  \\
\text{EWAC}^{\text{lb}} &:= \min_{\btheta \in \mathcal{F}} \ \ \text{EWAC}(\btheta).
\end{align*}
\end{subequations}
Denoting by $\text{EWAC}^*$ the true but generally unknown EWAC, we immediately obtain the following result.

\begin{restatable}[EWAC Bounds] {prop}{EWACBounds}
\label{prop:EWACBounds}
$
\text{EWAC}^{\text{lb}} \le \text{EWAC}^* \le \text{EWAC}^{\text{ub}}.
$
\end{restatable}

We pause here to consider why bounding the dishonest casino's EWAC only requires the solution of linear programs.  First, the EWAC is additive across time because each time period contributes a single term to the WAC (winnings attributable to cheating) and of course the expectation operator is additive. Taken together, this implies (see (\ref{eq:EWACDefn})) that we only need the marginal distribution of each counterfactual die roll $\tO_t$. Second, these marginal distributions are easy to compute because the action step involved a direct intervention on the hidden state, i.e., setting $\tH_t=\fair$.
It is these two features, i.e., the additivity of EWAC across time and the direct intervention on the hidden states, that results in the LP bounds. Indeed, the particular structure of the graphical model (HMM in our case) is of secondary importance. For example, we could still obtain LP bounds for the EWAC if we extended the HMM so that each $H_t$ depended directly on $(H_{t-1},O_{t-1})$ rather than just $H_{t-1}$, i.e., if we included an edge from each $O_{t-1}$ to $H_t$ in Figure \ref{fig:SCM-Casino1}. While $\delta_t(\fair)$ and $\delta_t(\loaded)$ would change, they could still be easily computed using standard message-passing algorithms. Moreover, a little consideration suggests that the proof of Proposition \ref{prop:EWACCharacterization} would go through unchanged.
It's quite remarkable then that we can bound a counterfactual query from the dishonest casino HMM via simple linear programs. Furthermore, in Section \ref{sec:CasinoExp}, we shall see how the explicit expression given by (\ref{eq:EWAC}) for any SCM's EWAC will enable us to  develop considerable intuition for the range of EWACs that are possible as well as the mechanisms giving rise to these EWACs.

\subsection{The Time-Inhomogeneous Case} \label{sec:TIH_LP}
We now consider time-{\em inhomogeneous} SCMs that are consistent with the HMM. While interesting in their own right, the resulting LPs will allow us to understand how much we can gain (in terms of tightness of the bounds) by imposing time-homogeneity.
We allow for time-inhomogeneity by simply allowing $\ftheta$ to be time-varying, and this leads to the following LP for the lower bound on EWAC:
\begin{subequations}
\label{eq:LPTime}
\begin{align}
    \min_{\ftheta^{1:T} } & \quad \wobs - \sum_{t=1}^T \left\{ \delta_t(\fair) \times w_{o_t} + \delta_t(\loaded) \times \sum_{i=1}^6 w_i \frac{\theta^{t}(i,o_t)}{e_{\loaded o_t}} \right\} \\
    \text{s.t.} & \quad \ftheta^t \in \mathcal{F} \ \ \forall t \in [T].
\end{align}
\end{subequations}
Clearly, we can obtain an upper bound by replacing the $\min$ with a $\max$ in (\ref{eq:LPTime}).
The LP in \eqref{eq:LPTime} is separable w.r.t.\ $t$ and this allows us to characterize its solution analytically. Omitting constants and scaling factors, the resulting LP for time $t$ is given by
\begin{align}
\label{eq:LPTimeDecomposed}
    \max_{\ftheta^t \in \mathcal{F}} \ \sum_{i=1}^6 w_i \, \theta^t(i,o_t).
\end{align}
(The minimization changes to a maximization due to the omission of the negative sign from the objective in (\ref{eq:LPTime}).) Interestingly, the optimal solution of this LP exhibits a closed-form solution.  To see why, suppose for the sake of illustration that $o_t=2$ and then consider the following 6-by-6 matrix of decision variables:
\begin{center}
\begin{tabular}{ c c c c c c | c }
 $\theta^t(1,1)$ & \blue{$\theta^t(1,2)$} & $\theta^t(1,3)$ & $\theta^t(1,4)$ & $\theta^t(1,5)$ & $\theta^t(1,6)$ & $e_{\fair  1}$ \\
 $\theta^t(2,1)$ & \blue{$\theta^t(2,2)$} & $\theta^t(2,3)$ & $\theta^t(2,4)$ & $\theta^t(2,5)$ & $\theta^t(2,6)$ & $e_{\fair  2}$ \\
 $\theta^t(3,1)$ & \blue{$\theta^t(3,2)$} & $\theta^t(3,3)$ & $\theta^t(3,4)$ & $\theta^t(3,5)$ & $\theta^t(3,6)$ & $e_{\fair  3}$ \\
 $\theta^t(4,1)$ & \blue{$\theta^t(4,2)$} & $\theta^t(4,3)$ & $\theta^t(4,4)$ & $\theta^t(4,5)$ & $\theta^t(4,6)$ & $e_{\fair  4}$ \\
 $\theta^t(5,1)$ & \blue{$\theta^t(5,2)$} & $\theta^t(5,3)$ & $\theta^t(5,4)$ & $\theta^t(5,5)$ & $\theta^t(5,6)$ & $e_{\fair  5}$ \\
 $\theta^t(6,1)$ & \blue{$\theta^t(6,2)$} & $\theta^t(6,3)$ & $\theta^t(6,4)$ & $\theta^t(6,5)$ & $\theta^t(6,6)$ & $e_{\fair  6}$ \\
 \hline
 $e_{\loaded 1}$ & $e_{\loaded 2}$ & $e_{\loaded 3}$ & $e_{\loaded 4}$ & $e_{\loaded 5}$ & $e_{\loaded 6}$ &
\end{tabular}
\end{center}
As dictated by $\mathcal{F}$, each row $i$ needs to sum to $e_{\fair  i}$ and each column $j$ to $e_{\loaded j}$. Since \eqref{eq:LPTimeDecomposed} maximizes $\sum_i w_i \, \theta^t(i,2)$ (recall $o_t = 2$), we focus on the second column which is highlighted in blue. Given that $w_1 < \ldots < w_6$, it is optimal to set $\theta^t(6,2)$ to be as large as possible and to proceed greedily so that
\begin{subequations}
\begin{align*}
\theta^t(6,2) &= \min\{e_{\fair  6}, e_{\loaded 2}\}  \\
\theta^t(5,2) &= \min\left\{e_{\fair  5}, e_{\loaded 2} - \theta^t(6,2)\right\} \\
\theta^t(4,2) &= \min\left\{e_{\fair  4}, e_{\loaded 2} - \theta^t(6,2) - \theta^t(5,2)\right\} \\
\vdots \nonumber \\
\theta^t(1,2) &= \min\left\{e_{\fair  1}, e_{\loaded 2} - \sum_{i > 1} \theta^t(i,2)\right\} .
\end{align*}
\end{subequations}
It is straightforward to generalize this pattern and we summarize this discussion in Proposition \ref{prop:CasinoTimeOpt}. (Note that the $\theta^t$ variables in other columns are ``free'' and can be set arbitrarily as long as the constraints in $\mathcal{F}$ are satisfied since they do not appear in the objective.  Also, one can characterize the upper bound solution similarly by proceeding in the reverse order, i.e., from row 1 to row 6.)

\begin{restatable}[Time-Inhomogenous Solution] {prop}{CasinoTimeOpt}
\label{prop:CasinoTimeOpt}
The optimal solution to \eqref{eq:LPTimeDecomposed} satisfies
\begin{align*}
    \theta^t(6,o_t) &= \min\{e_{\fair  6}, e_{\loaded o_t}\} \\
    \theta^t(i,o_t) &= \min\left\{e_{\fair  i}, \ e_{\loaded o_t} - \sum_{k > i} \theta^t(k,o_t)\right\}, \ i=5,\ldots , 1.
\end{align*}
\end{restatable}

It follows from Proposition \ref{prop:CasinoTimeOpt} that given $o_t$, the optimal period-$t$ solution $\ftheta^t$ is independent of $t$. Since  $o_t$ only takes values in $\{1, \ldots, 6\}$ ($\bO$ more generally), this means $\ftheta^t$ selects from only six ($|\bO|$ in general) distinct optimal values regardless of how large $T$ is. Furthermore,  unlike the time-homogeneous case, these optimal values are independent of the transition distributions $\fQ$ and only depend on the emission distributions $\fE$. We use this characterization in Appendix \ref{sec:AppTimeInHomo} to compare the time-homogeneous and time-inhomogeneous bounds.

\subsection{Partial Versus Full Identification of EWAC as $T \to \infty$} \label{sec:TtoInf}

The hidden Markov chain (under the dishonest policy) is a simple 2-state Markov chain that is both aperiodic and irreducible and therefore has a unique stationary distribution $\bpi \in \mathbb{R}^2$.  That an aperiodic and irreducible finite Markov chain has a unique stationary distribution is a well known result from the theory of Markov chains. See, for example, \cite{Ross1996}, which should be consulted for other Markov-chain related results that we use in this subsection.

The transition matrix for this 2-state Markov chain is
\begin{align} \label{eq:Q100}
\fQ &=
\begin{bmatrix}
\eta_1 & 1 - \eta_1 \\
\eta_2 & 1 - \eta_2
\end{bmatrix},
\end{align}
with the stationary distribution $\bpi = [\pi_1 \ \pi_2]$ being the unique solution to
\begin{equation} \label{ed:stat1}
\bpi^\top = \bpi^\top \fQ.
\end{equation}
It is easily seen that $\pi_1 = \eta_2/(1 + \eta_2 - \eta_1)$ and $\pi_2 = (1-\eta_1)/(1 + \eta_2 - \eta_1)$ solves (\ref{ed:stat1}). In the limit as $T \to \infty$, the fraction of time that the Markov chain will spend in the fair and biased states will converge to $\pi_1$ and $\pi_2$, respectively, with probability (w.p.) 1.

More generally, we can consider a {\em combined} 12-state Markov chain whose states consist of all possible hidden-state / emission combinations, i.e., all $(h,e)$ for $h \in \{\fair, \loaded\}$ and $e\in \{1, \ldots,6\}$. This is also an aperiodic and irreducible Markov chain and again has a unique stationary distribution $\bpi^{\mbox{\footnotesize c}} \in \mathbb{R}^{12}$. Rather than write out and solve a system of equations analogous to (\ref{ed:stat1}), it is clear from the structure of the HMM what form $\bpi^{\mbox{\footnotesize c}}$ takes. In particular, the stationary probability for states $(\fair,i)$ and $(\loaded,i)$ are simply $\pi_1 \times e_{\fair  i}$ and $\pi_2 \times e_{\loaded i}$, respectively, for $i \in \{1,\ldots,6\}$. Moreover, w.p.\ 1, these stationary probabilities are the long-run fractions of time that the Markov chain will spend in these states.

Hence, w.p.\ 1, the long-run time-average winnings of the casino under the dishonest policy is given by
\begin{eqnarray}
\overline{W}_{\mbox{\scriptsize dis}} &=& \pi_1 \, \sum_{i=1}^6 e_{\fair  i} w_i + \pi_2 \, \sum_{i=1}^6 e_{\loaded i} w_i \nonumber \nonumber \\
&=& \pi_1 \, \left(\sum_{i=1}^6 w_i\right)/6  + \pi_2 \, \sum_{i=1}^6 e_{\loaded i} w_i,  \label{eq:Stat2}
\end{eqnarray}
where we recall that the casino earns $w_i$ when a die roll of $i$ is observed. In contrast, under the no-cheating policy, it is clear that w.p.\ 1, the long-run time-average winnings of the casino is
\begin{equation}
\overline{W}_{\mbox{\scriptsize fair}} = \left(\sum_{i=1}^6 w_i\right)/6. \label{eq:Stat3}
\end{equation}
Hence, w.p.\ 1, the long-run time-average earnings due to cheating, i.e., average EWAC, is given by
\begin{subequations}
\begin{align}
\lim_{T \to \infty} \frac{\mbox{EWAC}}{T} &= \overline{W}_{\mbox{\scriptsize dis}} - \overline{W}_{\mbox{\scriptsize fair}} \label{eq:Stat44} \\
&= \pi_2 \, \left[\left(\sum_{i=1}^6 e_{\loaded i} w_i\right) -  \frac{\sum_{i=1}^6 w_i}{6}   \right], \label{eq:Stat4}
\end{align} \label{eq:Stat555}
\end{subequations}
where we have used (\ref{eq:Stat2}) and (\ref{eq:Stat3}) to substitute for $\overline{W}_{\mbox{\scriptsize dis}}$ and $\overline{W}_{\mbox{\scriptsize fair}}$, respectively.
We have an easy interpretation of (\ref{eq:Stat4}) as the average winnings when the biased die is thrown minus the average winnings when the fair die is thrown all weighted by $\pi_2$, the long run fraction of time the biased die is thrown.

What is interesting about expressions (\ref{eq:Stat2}) - (\ref{eq:Stat555}) is that they do not depend on the SCM / copula $\ftheta$. This is as expected since in the long run, i.e., as $T \to \infty$, we know w.p.\ 1 what the time-average earnings of the casino will be under either policy (cheating or no-cheating) and therefore know w.p.\ 1 what the time-average EWAC will be. In particular, this means that if we plot the lower and upper bounds of $\mbox{EWAC}/T$ against $T$ then we should see the bounds converge as $T \to \infty$.
In summary, $\mbox{EWAC}/T$ is only partially identifiable for any finite $T$ but it becomes perfectly identifiable in the limit as $T \to \infty$. We will demonstrate this in our numerical experiments in Section \ref{sec:NumTtoInf}.

\section{An Aside on Pathwise Monotonicity and Counterfactual Stability}  \label{sec:PM_CS}
Before proceeding to our numerical experiments, we briefly pause to consider the notions of pathwise monotonicity (PM) and counterfactual stability (CS), properties that are sometimes invoked to further constrain the space of feasible SCMs. We shall show that enforcing these properties is easy to do via linear constraints and therefore, does not create any additional difficulty in bounding the EWAC. If PM and / or CS are deemed appropriate for some components of the SCM, then (as we shall see later in our numerical experiments) they can often lead to much tighter bounds. In Section \ref{sec:PM}, we discuss PM and we do the same for CS in Section \ref{sec:CS}.

\subsection{Pathwise Monotonicity} \label{sec:PM}
Pathwise monotonicity or monotonicity  \citep[e.g.,][]{pearl_2009} is a simple and often intuitively appealing property that is often invoked in SCMs. In a medical context, for example, it simply states that a counterfactual outcome should be no worse than the outcome that actually occurred if the counterfactual intervention / treatment was better than what was actually applied.

\subsubsection*{Imposing PM for the Dishonest Casino}

If we want to impose PM in the dishonest casino setting, we can take either the casino's perspective or the casino's customers' perspective as both lead to the same set of constraints.  Taking the casino's perspective, imposing PM amounts to saying that for all $i=1,\ldots , 6$, if we obtain a die roll of $i$ under the fair die, then we should obtain a die roll of at least $i$ under the loaded die, i.e.,
\begin{equation} \label{eq:PM_Casino}
\theta(i,j) =0 \ \forall \, j < i.
\end{equation}
(Recall the casino wins more for higher die rolls so switching from the fair die to the loaded die is a better ``treatment'' for the casino and therefore makes the casino no worse off under pathwise monotonicity.) These are linear constraints and so, if PM is deemed appropriate, they are easily added to the constraint set $\mathcal{F}$ of our linear programming problems and will result in tighter bounds on the EWAC.
(For a physical casino with physical fair and loaded dice, we might think that an independent copula (so that $\theta(i,j)=e_{\fair  i}e_{\loaded j}$) is a feasible (or even likely) mechanism and this would rule out PM. For an online casino, however,  the true SCM will depend on the algorithm used to generate the die rolls and in principle, any SCM might be possible. However, as Remark \ref{comment:SimuateDieComon} at the end of Section \ref{sec:CasinoSCM} suggests, the comonotonic copula is the likely mechanism in this case and it is easy to see that the comonotonic copula implies PM. Of course, if we knew the comonotonic copula / SCM was used, then we could simply compute the true EWAC and would not need to solve LPs to bound it.)

\subsection{Counterfactual Stability} \label{sec:CS}
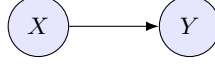
\begin{figure}[ht]
\centering
\begin{tikzpicture}[scale=1,transform shape]
    \node[state,fill=blue!10!white] (X) at (0,0) {\footnotesize $X$};
    \node[state,fill=blue!10!white] (Y) at (2,0) {\footnotesize $Y$};
    \path (X) edge (Y);
\end{tikzpicture}
\caption{A simple causal graph to illustrate CS.}
\label{fig:CS}
\end{figure}

Counterfactual stability (CS) has recently been proposed  \citep{ICML209-Obers} for handling counterfactual queries in certain settings. We can explain CS via the simple graphical model of Figure \ref{fig:CS}.
Suppose we observe an outcome $Y=y$ under policy $X=x$.
With  $Y_x := Y \mid (X = x)$,
CS requires that the counterfactual outcome under an interventional policy $\tx$ (denoted by $\tY := Y_{\tx} \mid Y_x = y$) cannot be $y'$ (for $y' \not = y$) if $\bP(Y_{\tx}=y) / \bP(Y_x=y) \geq \bP(Y_{\tx}=y') / \bP(Y_x=y')$.  In words, CS states that if $y$ was observed and this outcome becomes relatively more likely than $y'$ under the intervention, then the counterfactual outcome $\tY$ can not be $y'$.

Though somewhat appealing,  the appropriateness of CS depends on the application and should (like PM) be justified by domain-specific knowledge. For the most part, CS has been imposed via the Gumbel-max mechanism, see \cite{ICML209-Obers,NEURIPS2021_Tsirtsis} and indeed the former conjectured that the Gumbel-max mechanism was the unique mechanism / SCM that satisfies CS. \cite{HaughSingal2024}, however, showed that CS could be imposed via linear constraints in their polynomial programs and the non-tightness of their numerical bounds implied  the Gumbel-max mechanism did not uniquely satisfy CS.

\subsubsection*{Imposing CS for the Dishonest Casino}

In the context of the dishonest casino, CS can be imposed as follows.  With $e_{hi} = \Pb(O_{th} = i)$ denoting the probability that an observation equals $i$ given the hidden state is $h$, suppose that for arbitrary die roll observations $i \neq j$, we have
\begin{align*}
    \frac{e_{\fair  j}}{e_{\loaded j}} \ge \frac{e_{\fair i}}{e_{\loaded i}}.
\end{align*}
That is, observation $j$ is relatively more likely than $i$ under a fair die than under a loaded die. Then, CS requires
\[
0 = \Pb(O_{t\fair} = i, O_{t\loaded} = j) = \theta(i,j).
\]
But since $e_{\fair  i}=1/6$ for all $i$, we can therefore characterize the \emph{entire} space of SCMs that obey CS via the following linear constraints:
\begin{align}
\theta(i,j) = 0 \text{ for all } (i,j) \text{ such that } i \neq j \text{ and } e_{\loaded j} \le e_{\loaded i}. \label{eq:CS_HMM1}
\end{align}
Let $\text{EWAC}^{\text{ub}}_{\text{cs}}$ and $\text{EWAC}^{\text{lb}}_{\text{cs}}$ denote the upper and lower bounds we obtain when we add the CS constraints (\ref{eq:CS_HMM1}) to our constraint set $\mathcal{F}$. Then, the following result is immediate.
(Of course, the corresponding inequalities apply for the LPs where we impose the PM constraints.)
\begin{restatable}[EWAC CS Bounds] {prop}{EWACBoundsCS} \label{prop:EWACBoundsCS}
$
\text{EWAC}^{\text{lb}} \le \text{EWAC}^{\text{lb}}_{\text{cs}} \le \text{EWAC}^{\text{ub}}_{\text{cs}} \le \text{EWAC}^{\text{ub}}.
$
\end{restatable}
\noindent
Recalling that $\text{EWAC}^*$ denotes the EWAC under the true (unknown) SCM, we can only conclude that $\text{EWAC}^{\text{lb}}_{\text{cs}} \le \text{EWAC}^* \le \text{EWAC}^{\text{ub}}_{\text{cs}}$
{\em if} the true SCM satisfies CS.

\begin{rem} \label{rem:CSequalPM}
If the emission probability for the loaded die, i.e., $e_{\loaded i}$, is increasing in $i$, then (\ref{eq:CS_HMM1}) reduces to $\theta(i,j) = 0$ for all $(i,j)$ such that $i > j$. These constraints are then identical to the PM constraints in (\ref{eq:PM_Casino}).
\end{rem}

\section{Characterization of EWAC Under Benchmark SCMs} \label{sec:CopulaEGS}

In this brief section, we use three important benchmark SCMs to characterize possible values of the EWAC. These SCMs can be easily derived from the comonotonic, countermonotonic and independence copulas for $\ftheta$. These copulas are well understood and model extreme positive dependency, extreme negative dependency, and independence, respectively. As we shall see in our numerical experiments of Section \ref{sec:CasinoExp}, these benchmark SCMs prove to be very useful for developing intuition regarding the EWAC bounds. The following proposition states the benchmark EWAC values and its proof, together with a brief introduction to copulas, is provided in Appendix \ref{sec:CopulaAppendix}.

\begin{restatable}[{\bf EWAC Characterization Under Copulas}]{prop}{EWACCopulas} \label{prop:EWACCopulas}
Let $\EWACI$, $\EWACP$ and $\EWACN$ denote the EWAC for each of the independence, comonotonic and countermonotonic SCMs, respectively. Then,
\begin{align*}
\EWACI &= \wobs - \sum_{t=1}^T \left\{ \delta_t(\fair) w_{o_t} + \delta_t(\loaded) \sum_{i=1}^6 w_i e_{\fair  i}  \right\} \tag*{(independence)} \\
\EWACP &= \wobs - \sum_{t=1}^T \left\{ \delta_t(\fair) w_{o_t} + \delta_t(\loaded) \sum_{i=1}^6 w_i \frac{\thetaP(i,o_t)}{e_{\loaded o_t}} \right\} \tag*{(comonotonic)} \\
\EWACN &= \wobs - \sum_{t=1}^T \left\{ \delta_t(\fair) w_{o_t} + \delta_t(\loaded) \sum_{i=1}^6 w_i \frac{\thetaN(i,o_t)}{e_{\loaded o_t}} \right\} \tag*{(countermonotonic)}
\end{align*}
where, for all $(i,j)$,
\begin{align*}
\thetaP(i,j) &:= \sum_{\ell=0}^1 \sum_{\ell' = 0}^1 (-1)^{\ell + \ell'} \min\{\Theta_1(i-\ell),  \Theta_2(j-\ell') \} \\
\thetaN(i,j) &:= \sum_{\ell=0}^1 \sum_{\ell' = 0}^1 (-1)^{\ell + \ell'} (\Theta_1(i-\ell) + \Theta_2(j-\ell') - 1)^+,
\end{align*}
and for $h \in \{\fair, \loaded\}$, $\Theta_{h}(\cdot)$ is the CDF of $O_{th}$, i.e., $\Theta_h(i) = \Pb(O_t \le i \mid H_t = h) = \sum_{j \le i} e_{hj}$ for all $i$.
\end{restatable}

\section{Numerical Experiments}  \label{sec:CasinoExp}
We now describe our numerical experiments. We begin in Section \ref{sec:NumericalSetupCasino} with a description of our experimental setup and then describe our results for the time-homogeneous setting in Section \ref{sec:CasinoResults} and for the time-inhomogeneous setting in Section \ref{sec:Results_Time_Inhomo}.
In Section \ref{sec:NumTtoInf}, we demonstrate our analysis from Section \ref{sec:TtoInf}, i.e., that the time-average EWAC is identifiable in the limit as $T \to \infty$.
Some further numerical results on the nature of the solution to the time-inhomogeneous LP as well as {\em distribution} of WAC for given SCMs is provided in Appendix \ref{sec:AdditionalNums}.

\subsection{Problem Setup} \label{sec:NumericalSetupCasino}
We assume $T=30$ periods and assume $\eta_1=\eta_2 =: \eta$ in (\ref{eq:Q100}) so that $\eta \in [0,1]$ quantifies the degree of fairness in the HMM policy adopted by the casino. In particular, the initial state and transition distributions are given by
\begin{align*}
\fp &= (\eta, 1 - \eta) \nonumber \\
\fQ &=
\begin{bmatrix}
\eta & 1 - \eta \\
\eta & 1 - \eta
\end{bmatrix}.
\end{align*}
That is, the initial state, i.e.,  die, is fair w.p.\ $\eta$ and the next state is the same as the current state w.p.\ $\eta$ when the current die is fair and w.p.\ $1-\eta$ when the current die is biased. The emission distributions under the fair and loaded die obey
\begin{align*}
[e_{\fair  i}]_{i} \ &  \propto \ (1, \ldots, 1) \\
[e_{\loaded i}]_{i} \ &  \propto \ (1, \ldots, 6).
\end{align*}
Referring to Remark \ref{rem:CSequalPM}, we see that $e_{\loaded i}$ is increasing in $i$ and therefore the PM constraints coincide with the CS constraints: $\theta(i,j) = 0$ for all  $i > j$, i.e., $\Pb(O_{t\fair} > O_{t\loaded}) = 0$ for all $t$.
For this reason, we will only refer to the CS constraints in our main results below.
Finally, we set the casino's winnings to be $w_i := i$ for all $i$.
Recall that EWAC is defined given an observed path $o_{1:T}$ and we consider two paths:
\begin{eqnarray*}
\mbox{Path 1:} && (3, 5, 1, 2, 5, 4, 6, 3, 5, 2, 4, 3, 6, 4, 1, 2, 6, 4, 2, 3, 2, 1, 6, 3, 4, 1, 5, 1, 5, 6)\\
\mbox{Path 2:} && (6, 5, 6, 4, 1, 3, 5, 1, 2, 2, 6, 3, 4, 5, 5, 3, 2, 5, 6, 3, 4, 5, 5, 4, 6, 4, 4, 6, 5, 5).
\end{eqnarray*}
Path 1 has an average of 3.5 while Path 2 has an average of approximately 4.2. The first path therefore represents an ``unlucky'' path from the dishonest casino's perspective as its total winnings are precisely what it would have been expected to win under a policy of always using the fair die. The second path is more consistent with what might be expected under the HMM policy of occasionally using a loaded die.  For each path, we consider values of $\eta \in \{0.01, 0.02,  \ldots, 0.99\}$, with $\eta = 0$ and $\eta = 1$ denoting ``always cheating'' and ``never cheating'', respectively.

We coded in \texttt{MATLAB} \citep{MATLAB} and used \texttt{Gurobi} \citep{gurobi} to solve the LPs.  Our computations took a total of less than 10 minutes on a 3.8 GHz 8-Core Intel Core i7 processor with 16 GB 2667 MHz DDR4 memory. These 10 minutes include \emph{every} computation reported in the paper, i.e., all compute behind Figures \ref{fig:ewac1} to \ref{fig:ewac_dist}.
The code is available at the second author's website.

\subsection{The Time-Homogeneous Setting} \label{sec:CasinoResults}

\begin{figure}
  \begin{center}
    \subfigure[UB / LB]{\label{fig:ewac1a}\includegraphics[width=.45\linewidth]{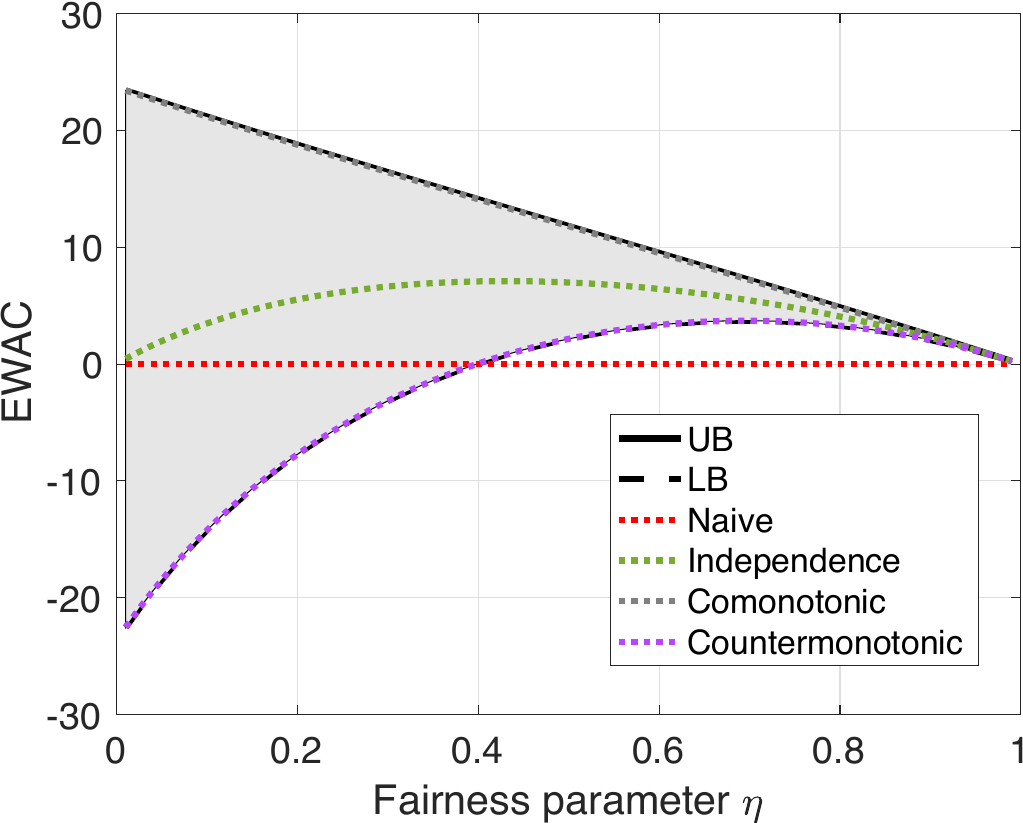}}
    \quad
    \subfigure[UB / LB with CS]{\label{fig:ewac1b}\includegraphics[width=.45\linewidth]{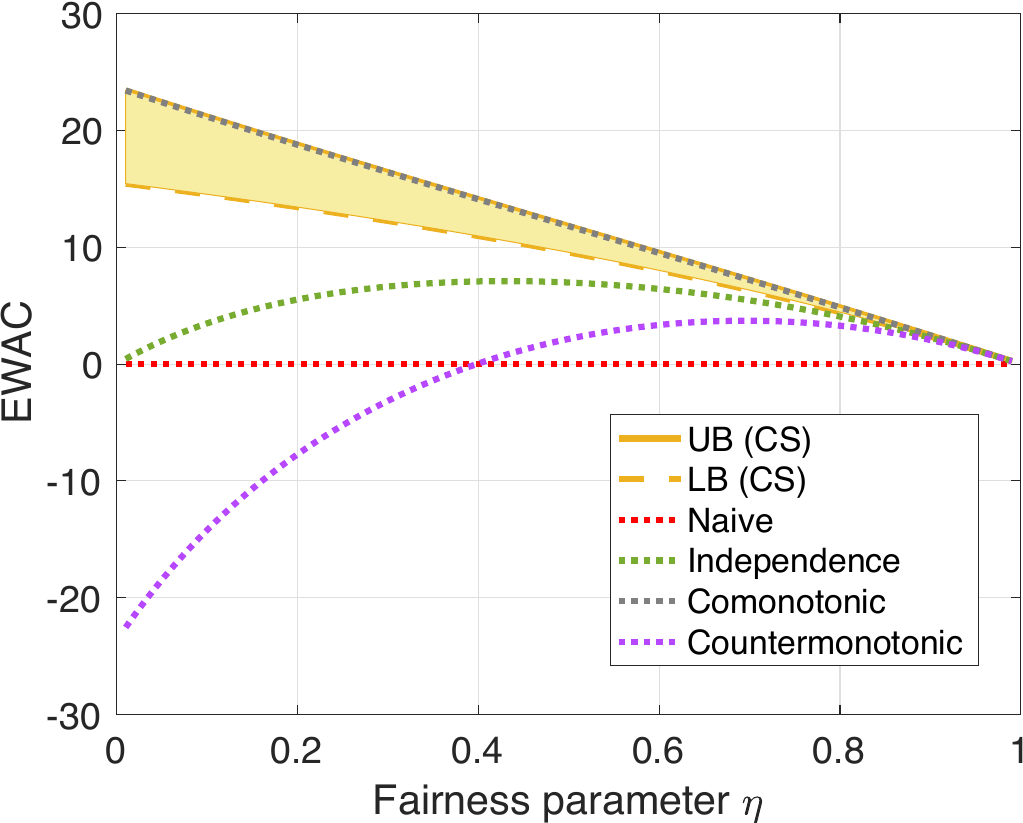}}
  \end{center}
\caption{EWAC results for Path 1.
In Figure \ref{fig:ewac1a}, the UB and comonotonic curves coincide (the highest curve in the figure), as do the LB and countermonotonic curves (the lowest curve).
In Figure \ref{fig:ewac1b}, the UB (CS) and comonotonic curves coincide (the highest curve in the figure). The region shaded gray in Figure \ref{fig:ewac1a} corresponds to feasible values of EWAC. The region shaded yellow in Figure \ref{fig:ewac1b} corresponds to feasible values of EWAC given that we impose the CS constraints. The independence and countermonotonic SCMs do not satisfy the CS constraints for any value of $\eta$.}
\label{fig:ewac1}
\end{figure}

\begin{figure}
  \begin{center}
    \subfigure[UB / LB]{\label{fig:ewac2a}\includegraphics[width=.45\linewidth]{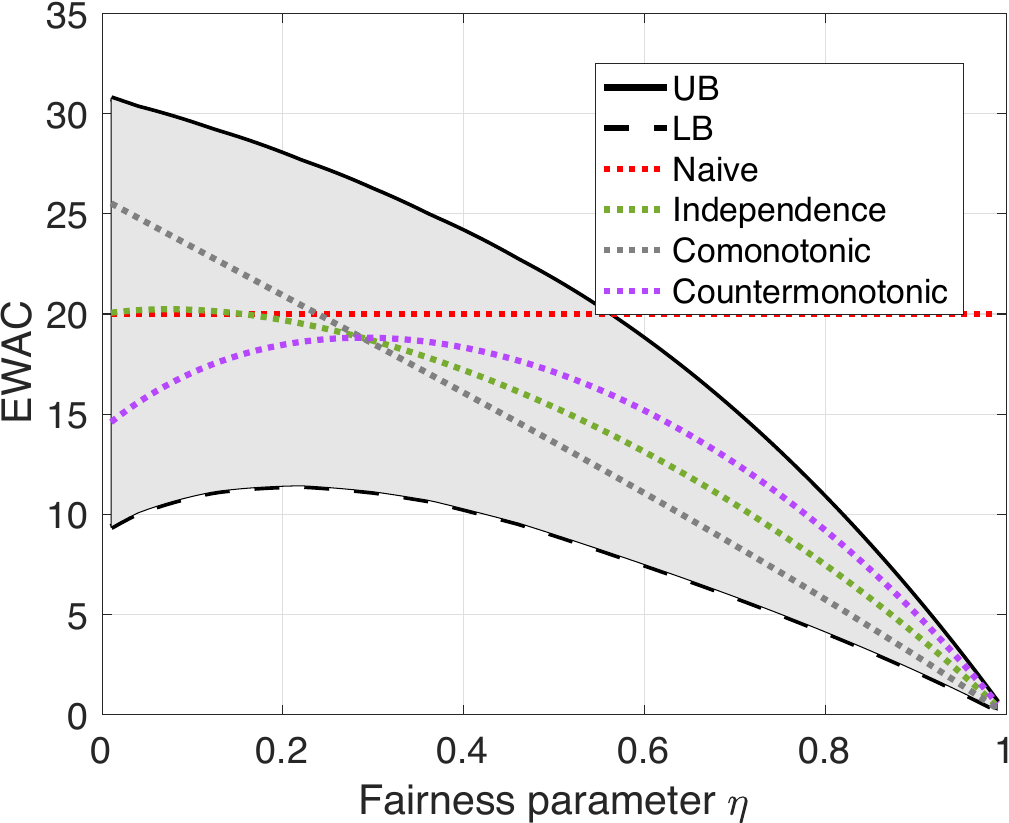}}
    \quad
    \subfigure[UB / LB with CS]{\label{fig:ewac2b}\includegraphics[width=.45\linewidth]{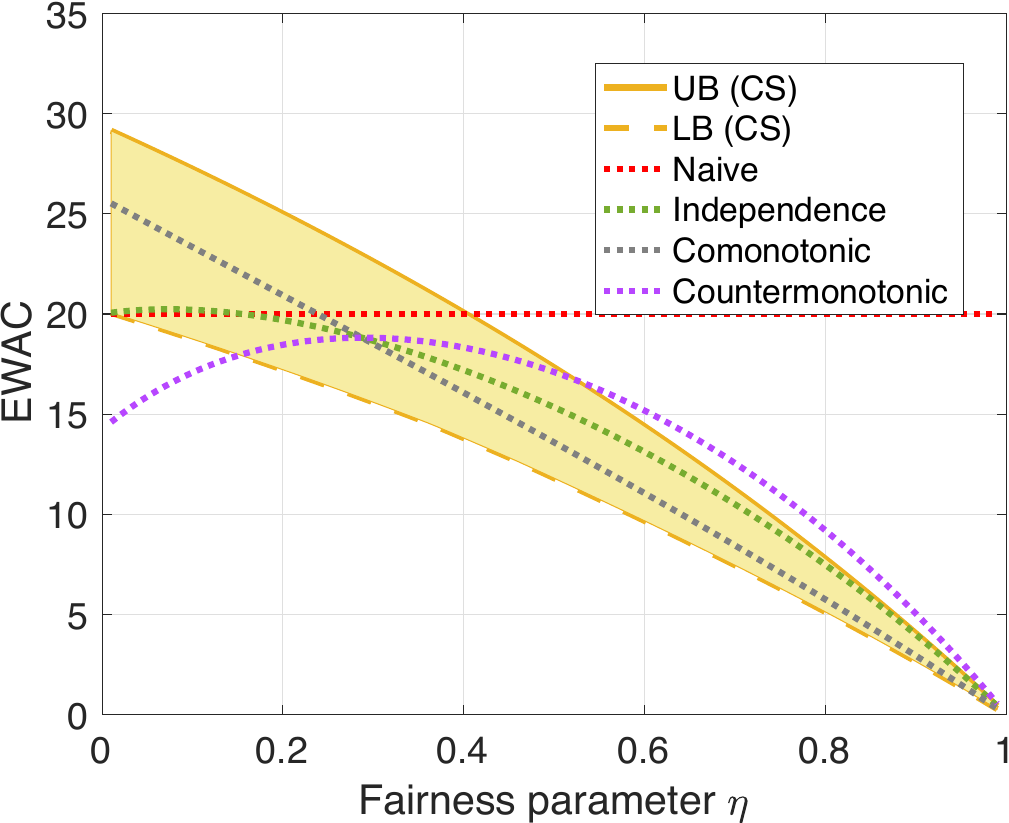}}
  \end{center}
\caption{EWAC results for Path 2. The region shaded gray in Figure \ref{fig:ewac1a} corresponds to feasible values of EWAC. The region shaded yellow in Figure \ref{fig:ewac1b} corresponds to feasible values of EWAC given that we impose the CS constraints. Only the countermonotonic SCM fails to satisfy the CS constraints for all values of $\eta$. }
\label{fig:ewac2}
\end{figure}

Figures \ref{fig:ewac1} and \ref{fig:ewac2} display the EWAC bounds (cf.\ Propositions \ref{prop:EWACBounds} and \ref{prop:EWACBoundsCS}) as a function of $\eta$ for the two paths. We also display the benchmark EWACs corresponding to the independence, comonotonic and countermonotonic SCMs (cf.\ Proposition \ref{prop:EWACCopulas}). Figures \ref{fig:ewac1a} and \ref{fig:ewac1b} are identical except in the latter, we impose the CS constraints and therefore obtain an often significant tightening of the bounds. (This is also true of Figures \ref{fig:ewac2a} and \ref{fig:ewac2b}.)
For example, when $\eta = 0.2$ in Figure \ref{fig:ewac1a}, the [LB,  UB] interval is approx.\ $[-8, 18]$ whereas the CS interval in Figure \ref{fig:ewac1b} is much tighter at approx.\ $[14, 18]$. We also note that every point in the CS interval corresponds to a copula / SCM that satisfies the CS property. As previously shown by \cite{HaughSingal2024},
this demonstrates that there are other causal mechanisms beyond the Gumbel-max mechanism that satisfy CS.

We also plot the naive estimate of EWAC that does not condition on the observed sequence of die rolls and therefore, completely ignores the abduction step. As discussed near the end of Section \ref{sec:CasinoOpt}, it simply calculates the expected winnings ($T\times 3.5=105$ since we assumed above $w_i=i$) if the casino were  to use a fair die on a {\em new} sequence of $T$ die rolls and subtracts this from the observed winnings. To interpret this naive EWAC,  consider Path 1 which corresponds to Figure \ref{fig:ewac1}. Path 1 is one where $\wobs = 105$ and so the naive estimate of EWAC is $105-105=0$ on this path. In contrast,  $\wobs = 125$ for Path 2 and so the naive estimate of EWAC on this path is $125 - 105 = 20$.  As we can see from Figure \ref{fig:ewac2} (corresponding to Path 2), the naive estimate of EWAC can lie outside the interval [LB,UB]. This just serves to emphasize that there is no causal mechanism that is consistent with the naive estimate. Moreover, the naive estimate does not depend on $\eta$ and is therefore constant in Figures \ref{fig:ewac1} and \ref{fig:ewac2}.

\subsubsection*{Further Discussion} \label{sec:FurtherDiscussion}
We now turn to $\EWACI$, the EWAC under the independence copula. For Path 1, we can see from Figure \ref{fig:ewac1} that $\EWACI$ starts at 0 ($\eta=0$) and ends at 0 ($\eta=1$), with a peak in between at approx.\ $\eta = 0.5$. This behavior can be explained via the $\EWACI$ characterization from Proposition \ref{prop:EWACCopulas}. In particular, we have
\begin{eqnarray}
\EWACI &=& \wobs - \sum_{t} \left\{ \delta_t(\fair) w_{o_t} + \delta_t(\loaded) \sum_{i} w_i e_{\fair  i} \right\} \nonumber \\
&=& \wobs - \sum_{t} \left\{ \delta_t(\fair) \times o_t + \delta_t(\loaded) \times 3.5 \right\} \label{eq:ExplainRes1}
\end{eqnarray}
since $e_{\fair  i} = 1/6$ for all $i$ and because we assumed $w_{i} = i$.
When $\eta = 0$, the casino always uses the loaded die and hence, $\delta_t(\fair)=0$ and $\delta_t(\loaded)=1$ for all $t$. Since $\wobs=105$ on this path, we obtain $\EWACI=0$. Similarly, when $\eta = 1$, the casino always uses the fair die and hence, $\delta_t(\fair)=1$ and $\delta_t(\loaded)=0$ for all $t$. In this case, (\ref{eq:ExplainRes1})
again yields  $\EWACI=0$ (since by definition $\wobs =  \sum_{t} w_{o_t} =  \sum_{t} o_t$).
For intermediate values of $\eta$, periods $t$ with a high value of $o_t$ will typically have a higher value of $\delta_t(\loaded)$ than periods with lower values of $o_t$. This is because the filtering / smoothing algorithm will generally infer that the loaded die is more likely to have been used when high die rolls, i.e.,  values of $o_t$, are observed. Referring to (\ref{eq:ExplainRes1}), this implies that more weight is placed on the 3.5 term than on the $o_t$ term when $o_t$ is high. Since the $o_t$'s have an average value of $3.5$ and $\sum_t o_t = \wobs$, this explains the peaked behavior of $\EWACI$ in Figure \ref{fig:ewac1} for intermediate values of $\eta$.
In the case of Path 2 and Figure \ref{fig:ewac2}, $\EWACI$ begins at 20 (when $\eta=0$) and monotonically decreases to 0 at $\eta=1$. This behavior can again be explained via (\ref{eq:ExplainRes1}). For example, when $\eta = 0$, $\delta_t(\fair)=0$ and $\delta_t(\loaded)=1$ for all $t$ and  this implies $\EWACI = \wobs - \sum_t 3.5 = 125 - 30 \times 3.5 = 20$. When  $\eta = 1$, $\delta_t(\fair)=1$ and $\delta_t(\loaded)=0$ for all $t$ and this implies $\EWACI = \wobs - \sum_t w_{o_t} = \wobs - \wobs = 0$.

The observation that all EWACs converge to 0 as $\eta \to 1$ can be explained via the general EWAC expression from Proposition \ref{prop:EWACCharacterization}:
\begin{align*}
\Scale[1]{\text{EWAC}(\ftheta) = \wobs - \sum_{t} \left\{ \delta_t(\fair) \times w_{o_t} + \delta_t(\loaded) \times \sum_{i} w_i \frac{\theta(i,o_t)}{e_{\loaded o_t}} \right\}.}
\end{align*}
When $\eta = 1$ (casino never uses the loaded die), we have $\delta_t(\fair)=1$ and $\delta_t(\loaded)=0$ for all $t$. Hence, $\text{EWAC}(\ftheta) = \wobs - \sum_t w_{o_t} = \wobs - \wobs = 0$ regardless of the causal mechanism / counterfactual joint distribution $\ftheta$.

We also observe that an EWAC can be negative. In Figure \ref{fig:ewac1}, for example,  the countermonotonic EWAC, i.e., $\EWACN$, is negative for small values of $\eta$. We can explain  this using the $\EWACN$ characterization in Proposition \ref{prop:EWACCopulas} but instead we will provide an intuitive explanation. As discussed earlier, when $\eta = 0$, we have $\delta_t(\fair)=0$ and $\delta_t(\loaded)=1$ for all $t$. This reflects the posterior certainty that the casino used a loaded die in every period. This in turn implies the casino should have made significantly more than $105$ in expectation. However, Path 1 corresponds to $\wobs = 105$, which implies the casino experienced a streak of ``bad luck'' despite always using the loaded die.
The countermonotonic copula flips the ``bad luck'' into ``good luck'' (compare the $U$ and $1-U$ in equation (\ref{eq:countermono})), and results in counterfactual winnings of over $105$. Subtracting these counterfactual winnings from $\wobs= 105$, we obtain a negative $\EWACN$. The same logic applies to non-zero but low values of $\eta$ on Path 1. Of course, if the countermonotonic curve is below 0, the LB has to be below 0 since by definition, LB is a lower bound (for \emph{all} feasible SCMs), and we clearly see this behavior for Path 1 in Figure \ref{fig:ewac1}.

In all of our results, we observe that $\EWACI$  lies between $\EWACP$ and $\EWACN$.
Under Path 1,  $\EWACP$ coincides with the $UB$ copula and $\EWACN$ coincides with the LB copula.
As demonstrated by Path 2 (Figure \ref{fig:ewac2}), this is not true in general, however, but they may serve as good approximations to the [LB, UB] range.
Furthermore,  we observe that $\EWACP$ obeys the CS property under both paths (since it always lies between the corresponding bounds) but $\EWACI$ and $\EWACN$ can violate the CS property, e.g., Path 1. Finally, we observe that even for a given path, the ordering among $\EWACI$, $\EWACP$ and $\EWACN$ can vary with $\eta$. This is clear from Figure \ref{fig:ewac2}.

\subsection{The Time-Inhomogeneous Setting} \label{sec:Results_Time_Inhomo}
\begin{figure}[ht]
  \begin{center}
    \subfigure[Path 1]{\label{fig:ewac1time}\includegraphics[width=.45\linewidth]{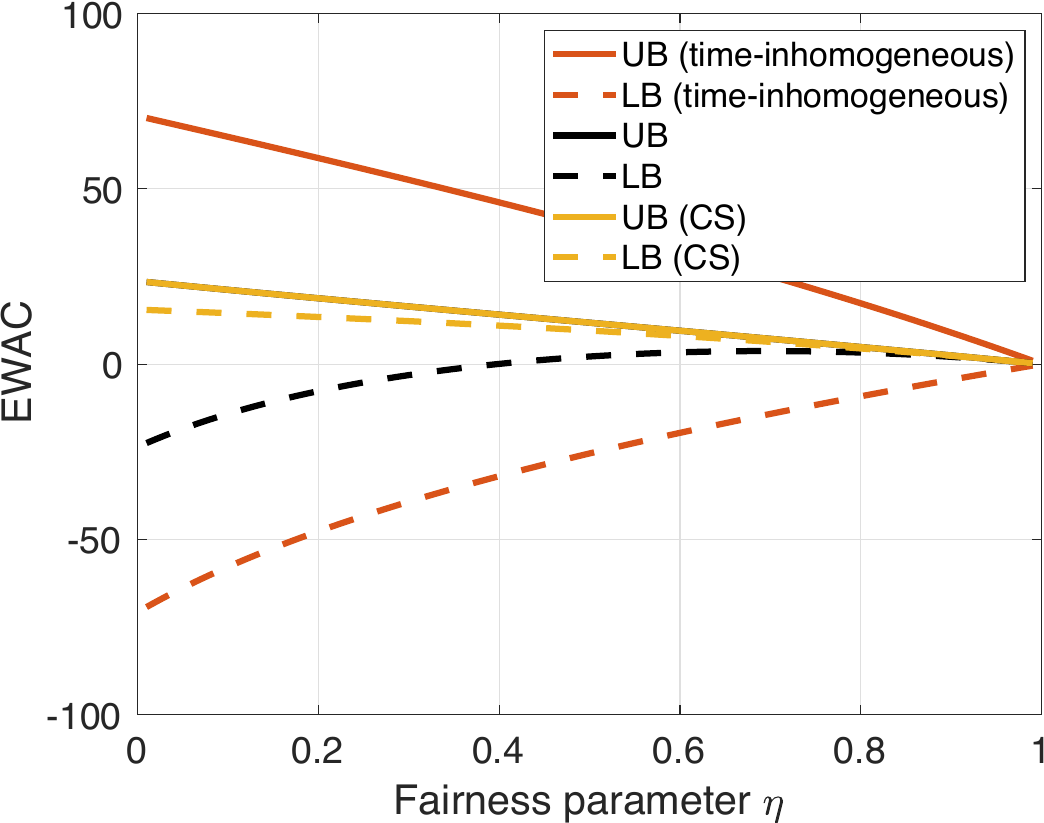}}
    \quad
    \subfigure[Path 2]{\label{fig:ewac2time}\includegraphics[width=.45\linewidth]{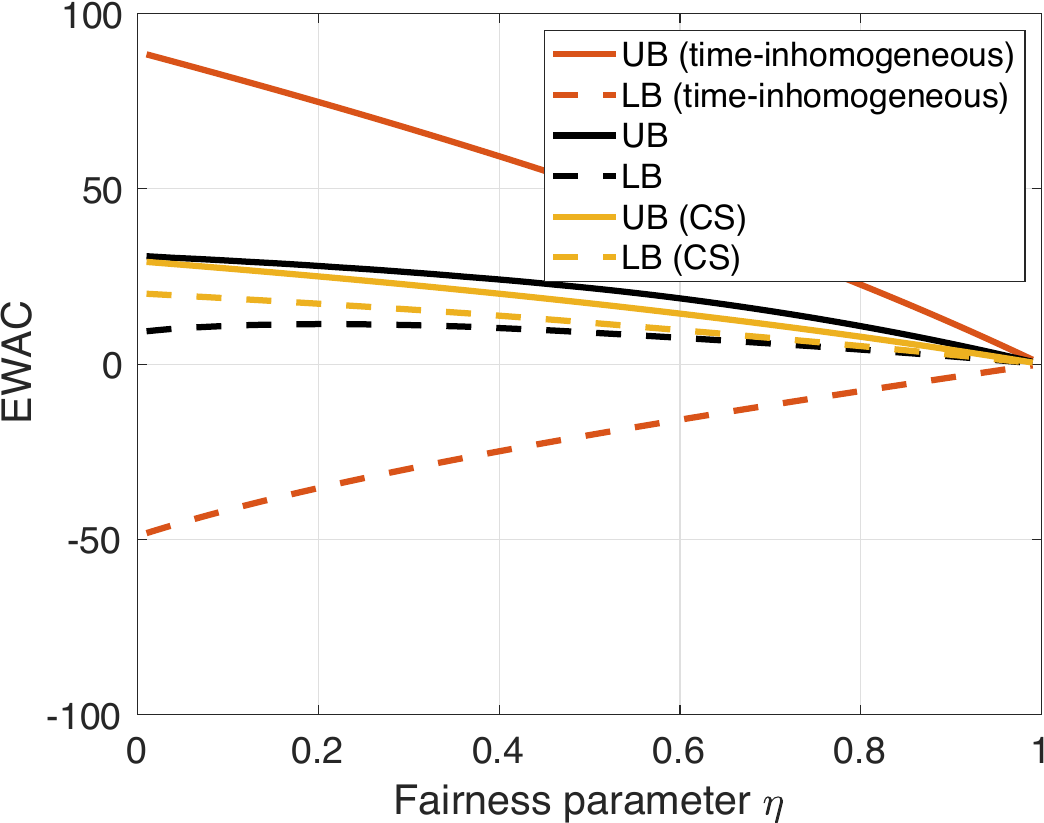}}
  \end{center}
\caption{EWAC bounds with time-homogeneous and time-inhomogeneous $\btheta$. The UB,  LB,  UB (CS) and LB (CS) all assume time-homogeneity of the SCM $\btheta$ and are therefore, identical to the corresponding curves in Figures \ref{fig:ewac1} and \ref{fig:ewac2}.
In Figure \ref{fig:ewac1time}, the UB and UB (CS) curves coincide. }
\label{fig:ewactime}
\end{figure}

Imposing time-homogeneity of the SCMs seems like a natural constraint to impose especially if the HMM dynamics are themselves time-homogeneous. However, as we saw in Section \ref{sec:TIH_LP}, we can easily bound the EWAC even when we allow for time-inhomogeneity of the SCM. Indeed, we saw that the resulting LP decouples into $T$ separate LPs which can be solved analytically as in Proposition \ref{prop:CasinoTimeOpt}. Figure \ref{fig:ewactime} displays the bounds for the time-inhomogeneous case on the same two paths that we considered in Section \ref{sec:CasinoResults}.
Clearly, there is a lot of value in imposing time-homogeneity in the sense that the time-homogeneous bounds are considerably closer than the time-inhomogeneous bounds. (We did not include the naive, independence, comonotonic, and countermonotonic EWACs in Figure \ref{fig:ewactime} in order to avoid cluttering the figures.)

In Appendix \ref{sec:AppTimeInHomo}, we study the solution to the time-inhomogeneous LP in further detail and discuss how it relates to the optimal solution to the time-inhomogeneous LP.

\subsection{Demonstrating the Identifiability of EWAC as $T \to \infty$}  \label{sec:NumTtoInf}

In this subsection, we demonstrate the results of Section \ref{sec:TtoInf}, namely that the time-average EWAC is identifiable in the limit as $T \to \infty$. In addition, as seen from (\ref{eq:Stat44}), this limiting time-average EWAC is simply the time-average of the naive EWAC we discussed in Section \ref{sec:CasinoResults}.
In these numerical experiments, we set $\eta = 0.5$ and varied $T$ between $10$ and $\Tmax:=10^5$ periods. We first sampled the observations path $o_{1:\Tmax}$ and then, for each $T$, we truncated $o_{1:\Tmax}$ to $o_{1:T}$ and compute the UB, LB, and the naive EWAC estimate using $o_{1:T}$. Note that when computing the $\text{EWAC}^{\text{ub}}$ and $\text{EWAC}^{\text{lb}}$ for a given $T$,  the delta terms, i.e., $[\delta_t(\fair), \delta_t(\loaded)]_{t=1}^T$, are computed using $o_{1:T}$ and not $o_{1:\Tmax}$. We then repeated this experiment two times, using a different random seed each time. We assumed the time-homogeneous setting when constructing $\text{EWAC}^{\text{ub}}$ and $\text{EWAC}^{\text{lb}}$ for each $T$ but clearly we could also have used the time-inhomogeneous setting.

\begin{figure}[H]
  \begin{center}
    \subfigure[Seed = 1]{\label{fig:ewac-inf-1-50}\includegraphics[width=.32\linewidth]{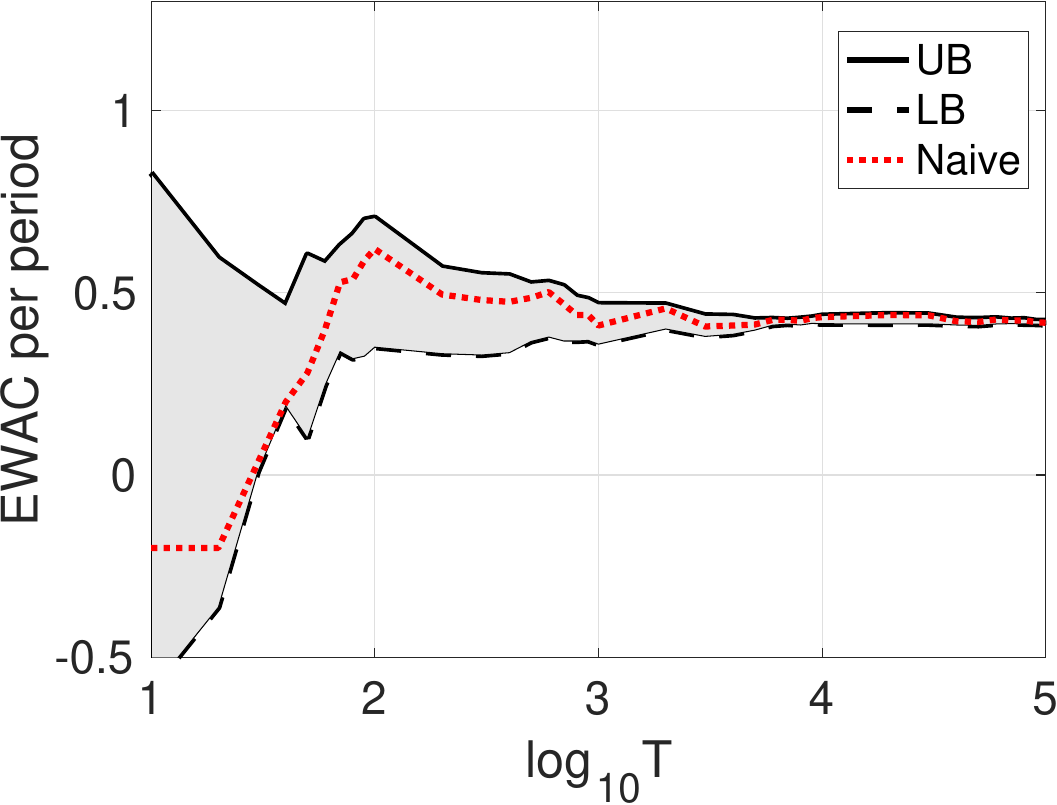}}
    \subfigure[Seed = 2]{\label{fig:ewac-inf-2-50}\includegraphics[width=.32\linewidth]{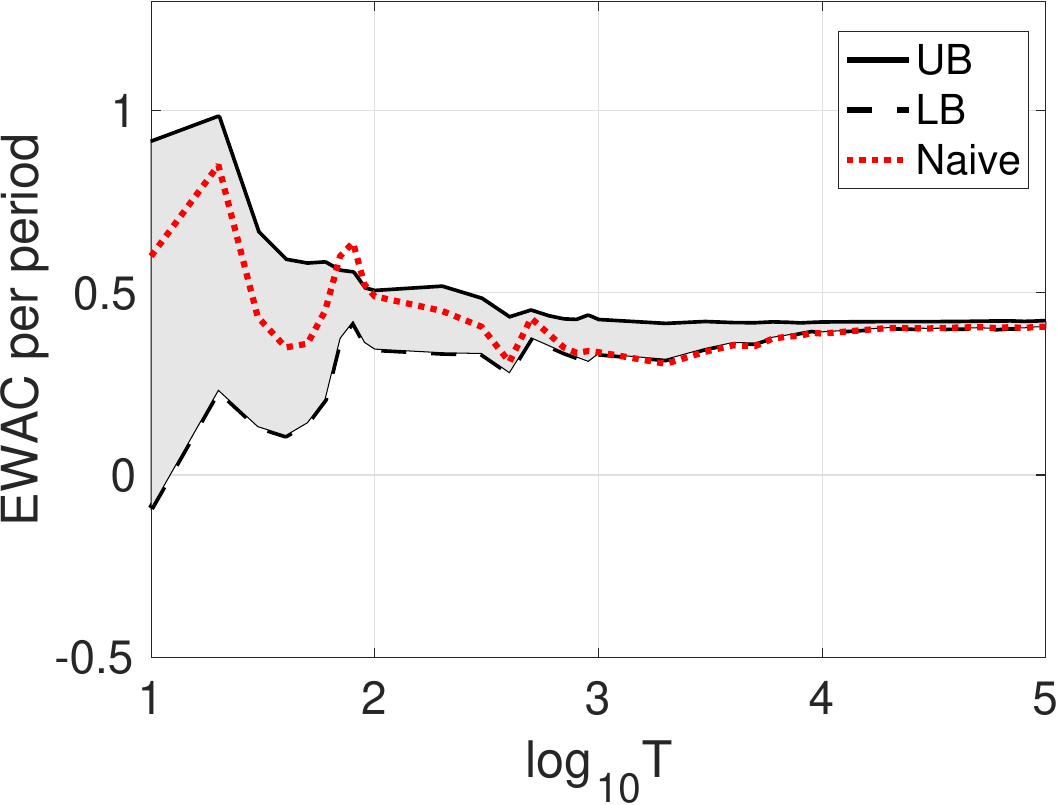}}
    \subfigure[Seed = 3]{\label{fig:ewac-inf-3-50}\includegraphics[width=.32\linewidth]{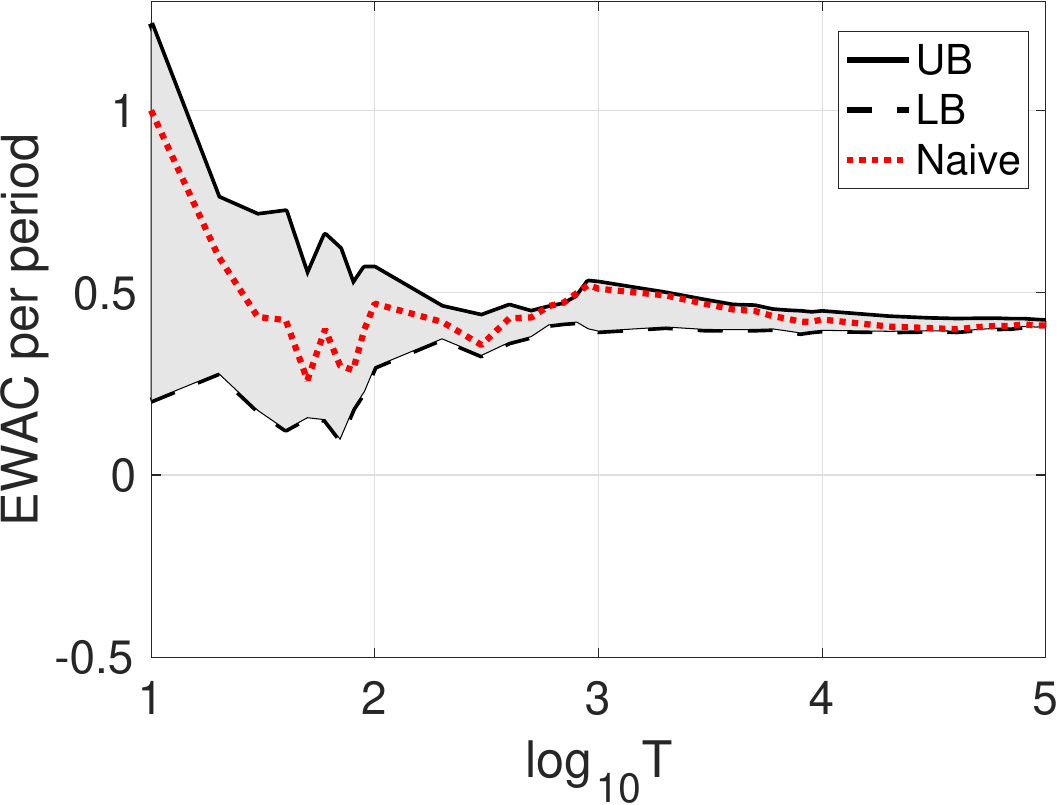}}
  \end{center}
\caption{Time-average EWAC as $T \to \infty$ with $\eta = 0.5$. Note the x-axis is on a log scale.}
\label{fig:ewac-inf}
\end{figure}

The results are displayed in Figure \ref{fig:ewac-inf} with the x-axis showing the number of periods $T$ on a log-scale. As expected, the time-average EWAC bounds converge to the time-average naive EWAC as $T \to \infty$. Moreover, the limiting value of the time-average naive EWAC does not depend on the seed, i.e., the realized path of die rolls, but of course the three paths (corresponding to the three different seeds) take different ``routes'' to converge to this shared limit.

\section{Concluding Remarks}  \label{sec:conc}
We have computed  lower and upper bounds for the dishonest casino's EWAC and utilize the explicit form of the EWAC in (\ref{eq:EWAC}) to help build intuition for just how varied the EWAC can be and what kinds of mechanisms, i.e., SCMs, can give rise to these bounds. The ability to bound a counterfactual query in a dynamic model via linear programs appears to be new. While it appears to have limited applications beyond the well-known dishonest casino setting, the explicit expression provided by (\ref{eq:EWAC}) should be useful in developing our general understanding of counterfactual modeling in HMMs and extensions of HMMs.  We also showed the average EWAC is identifiable in the limit as $T \to \infty$, which contrasts with the case of finite $T$ where the average EWAC is only partially identifiable.

There are several interesting directions for future research. One direction would be in developing extended HMM models where the tractability of the dishonest casino setting still prevails. For example, it may be possible to build such models in marketing or revenue management contexts where, for example, the $H_t$'s represent the unknown type of a customer and $w_{o_t}$ the revenue obtained from such a customer. \\

\paragraph{Acknowledgements} The authors are grateful to Jonas Lieber and Andre Veiga for helpful conversations. They are also grateful for the reviewers' valuable comments that improved the manuscript.

\paragraph{Funding Information} Authors state no funding involved. 

\paragraph{Author Contribution} All authors have accepted the responsibility for the content of the manuscript and consented to its submission, reviewed all the results, and approved the final version of the manuscript. RS developed the model code and performed the simulations.

\paragraph{Conflict of Interest} Authors state no conflict of interest. 

\paragraph{Data Availability Statement} The datasets generated during and/or analysed during the current study are available at \href{https://www.columbia.edu/~rs3566/files/codeCasino.zip}{\texttt{https://www.columbia.edu/$\sim$rs3566/files/codeCasino.zip}}.

\newpage

\bibliography{bibliography}

\clearpage
\begin{appendices}

\section{Using Copulas to Estimate EWAC}  \label{sec:CopulaAppendix}
In this appendix, we discuss the role of copulas in our counterfactual analysis and how specific copulas can be used to provide benchmark values of the dishonest casino's EWAC.
Copulas are functions that enable us to separate the marginal distributions from the dependency structure of a given multivariate distribution. They are particularly useful in applications where the marginal distributions are known  but a joint distribution with these known marginals is required.
This situation arises in many applications including both insurance and finance.  In finance for example, the market prices of options on individual securities or indices can be used to compute the so-called risk-neutral (marginal) distributions for these securities. But if one is pricing an option on a {\em basket} of individual securities, then the {\em joint} risk-neutral distribution is required. A similar situation arises with credit-default swaps (CDS). The market prices of CDS's can be used to infer the marginal risk-neutral probability of a company declaring bankruptcy by a certain date. But a collateralized debt obligation (CDO) depends on the {\em joint} risk-neutral distribution of the underlying companies going bankrupt.
In the dishonest casino setting of this paper, we know the univariate marginal distributions of $(O_{t\fair}, O_{t\loaded})$ as they are given by the appropriate emission distribution.

In each of these cases, one needs to work with a joint distribution with fixed or pre-specified marginal distributions. Copulas and Sklar's Theorem (see below) can be very helpful in these situations. We only briefly discuss the main definitions and results from the theory of copulas here but \cite{Nelsen2006} can be consulted for an introduction to the topic. \cite{QRM2015} also contains a nice introduction but in the context of financial risk management.

\begin{defi}[Copula] \label{def:Copula}
A $d$-dimensional {copula}, $C : [0,1]^d : \rightarrow [0,1]$ is a cumulative distribution function with uniform marginals.
\end{defi}

We write $C({\bf u}) = C(u_1, \ldots , u_d)$ for a generic copula. It follows immediately from Definition \ref{def:Copula} that  $C(u_1, \ldots , u_d)$ is non-decreasing in each argument and that $C(1,\ldots , 1, u_i, 1, \ldots , 1)  =  u_i$. It is also easy to confirm that $C(1,u_1, \ldots , u_{d-1})$ is a $(d-1)$-dimensional copula and, more generally, that all $k$-dimensional marginals with $2 \leq k \leq d$ are copulas.  The most important result from the theory of copulas is Sklar's Theorem \citep{Sklar1959}.

\begin{thm}[Sklar 1959]
Consider a $d$-dimensional CDF $\bPi$ with marginals $\bPi_1$, \ldots , $\bPi_d$. Then, there exists a copula $C$ such that
\begin{align}
\label{eq:Sklar1}
\bPi(x_1, \ldots , x_d) = C\left(\bPi_1(x_1), \ldots , \bPi_d(x_d)  \right)
\end{align}
for all $x_i \in [-\infty, \; \infty]$ and $i=1, \ldots, d.$

If $\bPi_i$ is continuous for all $i = 1, \ldots , d$, then $C$ is unique; otherwise $C$ is uniquely determined only on $\mbox{Ran}(\bPi_1) \times \cdots \times \mbox{Ran}(\bPi_d)$, where $\mbox{Ran}(\bPi_i)$ denotes the range of the CDF $\bPi_i$.

Conversely, consider a copula $C$ and univariate CDF's $\bPi_1, \ldots , \bPi_d$. Then, $\bPi$ as defined in (\ref{eq:Sklar1}) is a multivariate CDF with marginals $\bPi_1, \ldots , \bPi_d$.
\end{thm}

A particularly important aspect of Sklar's Theorem in the context of this paper is that $C$ is only uniquely determined on $\mbox{Ran}(\bPi_1) \times \cdots \times \mbox{Ran}(\bPi_d)$. Because we are interested in applications with discrete state-spaces, this implies that there will be many copulas that lead to the same joint distribution $\bPi$. It is for this reason that we prefer to work directly with the joint PMF of $(O_{t\fair},  O_{t\loaded})$ in Section \ref{sec:CasinoLPs} rather than the copula of $\fV_t$ in Section \ref{sec:CasinoSCM}. That said, we emphasize that specifying copulas for the exogenous vectors $\fU_t$ and $\fV_t$ is equivalent to specifying a particular structural causal model (SCM) in which the EWAC can be computed.
The following important result was derived independently by Fr\'{e}chet and Hoeffding and provides lower and upper bounds on copulas.

\begin{thm}[The Fr\'{e}chet-Hoeffding Bounds] Consider a copula $C({\bf u}) = C(u_1, \ldots , u_d)$. Then,
\begin{align*}
\max \left\{1 - d + \sum_{i=1}^d u_i,  0  \right \}  \leq  C({\bf u})  \leq \ \min \{u_1, \ldots , u_d \}.
\end{align*}
\end{thm}
We now define the three copulas that we use to define our benchmark SCMs.

\begin{defi}[Comonotonic Copula] \label{def:ComonotonicCopula}
The comonotonic copula is defined according to
\begin{align*}
\CP({\bf u}) := \min \{u_1, \ldots , u_d  \}
\end{align*}
which coincides with the Fr\'{e}chet-Hoeffding upper bound. It corresponds to the case of  extreme positive dependence. For example, let $\fU =(U_1, \ldots , U_d)$ with $U_1 = U_2 = \cdots = U_d  \sim \text{U}[0,1]$. Then, clearly $\min \{u_1, \ldots , u_d  \} = \bPi(u_1,\ldots , u_d)$ but by Sklar's Theorem $F(u_1,\ldots , u_d) = C(u_1,\ldots , u_d)$ and so, $ C(u_1,\ldots , u_d) = \min \{u_1, \ldots , u_d  \}$.
\end{defi}

\begin{defi}[Countermonotonic Copula] \label{def:CountermonotonicCopula}
The countermonotonic copula is a 2-dimensional copula given by
\begin{align}
\label{eq:CounterM1}
\CN({\bf u}) := \max \{u_1 + u_2 - 1,   0 \},
\end{align}
which coincides with the Fr\'{e}chet-Hoeffding lower bound when $d=2$. It corresponds to the case of extreme negative dependence.  It is easy to check that (\ref{eq:CounterM1}) is the joint distribution of $(U,1-U)$ where $U \sim \text{U}[0,1]$. (The Fr\'{e}chet-Hoeffding lower bound is only tight when $d=2$. This is analogous to the fact that while a pairwise correlation can lie anywhere in $[-1,1]$, the {\em average} pairwise correlation of $d$ random variables is bounded below by $-1/(d-1)$.)
\end{defi}

\begin{defi}[Independence Copula] \label{def:IndependenceCopula}
The independence copula satisfies
\begin{align*}
\CI({\bf u})  :=  \prod_{i=1}^d u_i,
\end{align*}
and it's easy to confirm using Sklar's Theorem that random variables are independent if and only if their copula is the independence copula.
\end{defi}

There are many other well-known classes of copulas including, for example, Archimedean, Gaussian and t copulas. It is also easy to check that convex combinations of copulas are copulas and so it is straightforward to create other benchmark SCMs. We are now ready to prove Proposition \ref{prop:EWACCopulas} which provides values of the casino's EWAC for each of the three SCMs defined by assuming the independence, comonotonic and countermonotonic copulas for $(O_{t\fair},  O_{t\loaded})$.

\EWACCopulas*
\paragraph{Proof}
It follows from (\ref{eq:EWAC}) that we simply need to characterize
$\theta(i,o_t)$ under each of the three copulas. For the independence copula, we have
\begin{align*}
\thetaI(i,j) = \Pb( O_{t\fair} = i,  O_{t\loaded} = j)
    &= \Pb( O_{t\fair} = i) \times \Pb(O_{t\loaded} = j) \\
    &= e_{\fair  i} \times e_{\loaded j}
\end{align*}
and we obtain $\EWACI$.
For the comonotonic and countermonotonic copulas,  the following general fact proves useful:
\begin{align}
\theta(i,j) &= \Pb(O_{t\fair} = i, O_{t\loaded} = j) \nonumber \\
&= \Pb(O_{t\fair} \le i, O_{t\loaded} \le j) - \Pb(O_{t\fair} < i, O_{t\loaded} \le j) - \Pb(O_{t\fair} \le i, O_{t\loaded} < j) + \Pb(O_{t\fair} < i, O_{t\loaded} < j) \nonumber \\
&= \Theta(i,j) - \Theta(i-1,j) - \Theta(i,j-1) + \Theta(i-1,j-1) \nonumber \\
&= \sum_{\ell=0}^1 \sum_{\ell'=0}^1 (-1)^{\ell + \ell'} \Theta(i-\ell, j-\ell') \label{eq:Pitopi}
\end{align}
where $\Theta(i,j) = \Pb(O_{t\fair} \le i, O_{t\loaded} \le j)$ is the joint counterfactual CDF. Recalling that $\Theta_h(i) = \Pb(O_{th} \le i)$ for all $(h,i)$, the comonotonic joint CDF $\ThetaP$ satisfies
\begin{subequations}
\label{eq:CoCopula}
\begin{align}
\ThetaP(i,j) &= \Pb(O_{t\fair} \le i, O_{t\loaded} \le j) \nonumber \\
   &= \Pb(\Theta_1^{-1}(U) \le i, \Theta_2^{-1}(U) \le j) \label{eq:comono} \\
   &= \Pb(U \le \Theta_1(i),  U \le \Theta_2(j)) \nonumber \\
   &= \Pb(U \le \min\{\Theta_1(i),  \Theta_2(j)\}) \nonumber \\
   &= \min\{\Theta_1(i),  \Theta_2(j) \} \label{eq:comon1}
\end{align}
\end{subequations}
where comonotonicity is invoked in \eqref{eq:comono} with $U \sim \text{U}[0,1]$. Substituting \eqref{eq:comon1} into \eqref{eq:Pitopi} yields
\begin{align*}
\thetaP(i,j) = \sum_{\ell=0}^1 \sum_{\ell' = 0}^1 (-1)^{\ell + \ell'} \min\{\Theta_1(i-\ell),  \Theta_2(j-\ell') \}
\end{align*}
for all $(i,j)$.
Finally, the countermonotonic joint counterfactual CDF satisfies
\begin{subequations}
\label{eq:CounterCopula}
\begin{align}
\ThetaN(i,j) &= \Pb(O_{t\fair} \le i, O_{t\loaded} \le j) \nonumber \\
   &= \Pb(\Theta_1^{-1}(U) \le i, \Theta_2^{-1}(1-U) \le j) \label{eq:countermono} \\
&= \Pb(U \le \Theta_1(i),  1-U \le \Theta_2(j)) \nonumber \\
   &= \Pb(U \le \Theta_1(i),  U \ge 1 - \Theta_2(j)) \nonumber \\
   &= \Pb(U \in [ 1 - \Theta_2(j),  \Theta_1(i) ]) \label{eq:countermono1} \\
   &= (\Theta_1(i) + \Theta_2(j) - 1)^+ \nonumber
\end{align}
\end{subequations}
where countermonotonicity is invoked in \eqref{eq:countermono} with $U \sim \text{U}(0,1)$, and $(\cdot)^+ := \max\{\cdot, 0\}$.
Substituting \eqref{eq:countermono1} into \eqref{eq:Pitopi} yields
\begin{align*}
\thetaN(i,j) = \sum_{\ell=0}^1 \sum_{\ell' = 0}^1 (-1)^{\ell + \ell'} (\Theta_1(i-\ell) + \Theta_2(j-\ell') - 1)^+
\end{align*}
for all $(i,j)$. \hfill $\Box$

\begin{rem}
The non-uniqueness of the copula $C$ for discrete random variables implies the existence of other copulas that will also lead to the same values of $\EWACI$, $\EWACP$ and $\EWACN$. Nonetheless, we refer to them as the independence, comonotonic and countermonotonic values of EWAC.
\end{rem}

\section{Additional Numerical Results} \label{sec:AdditionalNums}
In this appendix, we provide some additional insight and results. In Section \ref{sec:AppTimeInHomo}, we discuss the nature of the optimal solution for the time-inhomogeneous LP and how it relates to the optimal solution for the time-homogeneous LP. Then, in Section \ref{sec:DistributionWAC}, we consider the distribution of the WAC.

\subsection{The Solution to the Time-Inhomogeneous LP}  \label{sec:AppTimeInHomo}

To further understand why the time-inhomogeneous bounds are wider than the time-homogeneous ones,  consider the LB (identical intuition holds for the UB as well). The same numerical setup of Section \ref{sec:CasinoExp} (\( e_{\fair i} = 1/6 \) for all $i$ and \( e_{\loaded j} = j/21 \) for all \( j \)), combined with Proposition \ref{prop:CasinoTimeOpt}, implies that the optimal \( \ftheta^t \) (as a function of \( o_t \)) is given by:
\begingroup
\small
\setlength{\arraycolsep}{3pt}
\begin{align*}
\ftheta^t \mid (o_t = 1) &=
\begin{bmatrix}
 & & & & & \\ & & & & & \\ & & & & & \\ & & & & & \\ & & & & & \\ 1/21 & & & & &
\end{bmatrix}, \quad
\ftheta^t \mid (o_t = 2) =
\begin{bmatrix}
 & & & & & \\ & & & & & \\ & & & & & \\ & & & & & \\ & & & & & \\ & 2/21 & & & &
\end{bmatrix},  \quad
\ftheta^t \mid (o_t = 3) =
\begin{bmatrix}
 & & & & & \\ & & & & & \\ & & & & & \\ & & & & & \\ & & & & & \\ & & 3/21 & & &
\end{bmatrix},  \\
\ftheta^t \mid (o_t = 4) &=
\begin{bmatrix}
 & & & & & \\ & & & & & \\ & & & & & \\ & & & & & \\ & & & 1/42 & &  \\ & & & 1/6 & &
\end{bmatrix},  \quad
\ftheta^t \mid (o_t = 5) =
\begin{bmatrix}
 & & & & & \\ & & & & & \\ & & & & & \\ & & & & & \\ & & & & 3/42 & \\ & & & & 1/6 &
\end{bmatrix}, \quad
\ftheta^t \mid (o_t = 6) =
\begin{bmatrix}
 & & & & & \\ & & & & & \\ & & & & & \\ & & & & & \\ & & & & & 5/42 \\ & & & & & 1/6
\end{bmatrix}.
\end{align*}
\endgroup
As discussed around Proposition \ref{prop:CasinoTimeOpt}, the non-filled entries remain ``free'' as long as the constraints in \( \cF \) are satisfied, meaning each row \( i \) sums to \( e_{\fair i} \) (\( =1/6 \) here) and each column \( j \) sums to \( e_{\loaded j} \) (\( =j/21 \) here). Combining the six solutions into a single matrix, we obtain:
\begin{align*}
{\bf A} :=
\begin{bmatrix}
 & & & & & \\
 & & & & & \\
 & & & & & \\
 & & & & & \\
 & & & 1/42 & 3/42 & 5/42 \\
1/21 & 2/21 & 3/21 & 1/6 & 1/6 & 1/6
\end{bmatrix}.
\end{align*}
Trivially, if ${\bf A} $ satisfied the constraints in \( \cF \), the optimal time-homogeneous solution, $\btheta^{\text{lb}}$ say, would satisfy $\btheta^{\text{lb}} = {\bf A}$, making the time-homogeneous LB equal to the time-inhomogeneous LB. However, this is not the case. Specifically, while each column \( j \) sums to \( j/21 \), rows 5 and 6 exceed \( 1/6 \), resulting in a lower time-inhomogeneous LB compared to the time-homogeneous LB.

\subsection{Distribution of Winnings Attributable to Cheating (WAC)}   \label{sec:DistributionWAC}
While we have focused on computing or bounding the \emph{expected} WAC, i.e., the EWAC, in the main body of the paper, it is also possible to understand the \emph{distribution} of WAC for a given SCM by simulating from the full joint distribution of $\tO_{1:T}$.
Towards this end, consider an arbitrary joint PMF $\ftheta$ for $(O_{t\fair}, O_{t\loaded})$ as defined in Section \ref{sec:CasinoLPs}. This could be $\thetaI$, $\thetaP$, or $\thetaN$ from the proof of Proposition \ref{prop:EWACCopulas} in Appendix \ref{sec:CopulaAppendix}, for example, or indeed the PMF corresponding to any of the EWAC bounds.  Our goal is to understand the distribution of WAC under the given $\ftheta$ conditional of course on the observed path $o_{1:T}$.  To do so, we first generate $S$ posterior samples of the hidden path $H_{1:T} \mid o_{1:T}$ via the FFBS algorithm \citep{barber2012bayesian}. We denote these simulated paths by $[h_{1:T}(s)]_{s=1}^S$.   Second, for each sampled path $h_{1:T}(s)$, we loop over each hidden state $h_{t}(s)$ for $t \in [T]$. If the hidden state is ``fair'', i.e., $h_{t}(s) = \fair$, then we have $\tildeo_t(s) = o_t$. Otherwise, $h_{t}(s) = \loaded$ and we use the joint distribution $\ftheta$ conditioned on the observation $o_t$ and $H_t=\loaded$ to sample $\tildeo_t(s)$. (After generating the hidden path $h_{1:T}(s)$, the steps are identical to those carried out in the proof of Proposition \ref{prop:EWACCharacterization} except that we know each hidden state $h_t(s)$ and therefore do not need to weigh by the posterior probabilities $\delta_t(\fair)$ and $\delta_t(\loaded)$.) For each path $h_{1:T}(s)$, we then have a corresponding path $\tildeo_{1:T}(s)$ and a corresponding WAC.

\begin{figure}[ht]
  \begin{center}
    \subfigure[Path 1]{\label{fig:ewac_dist1}\includegraphics[width=.48\linewidth]{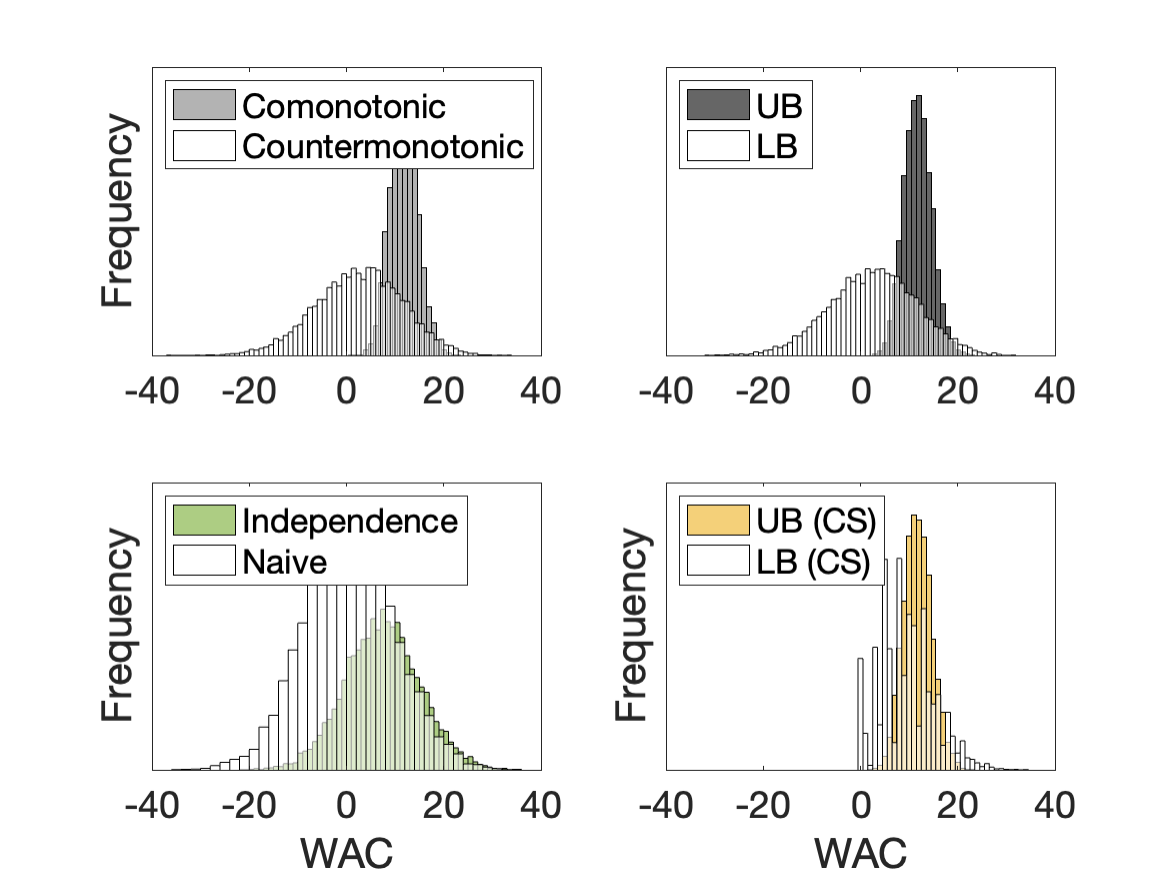}}
    \quad
    \subfigure[Path 2]{\label{fig:ewac_dist2}\includegraphics[width=.48\linewidth]{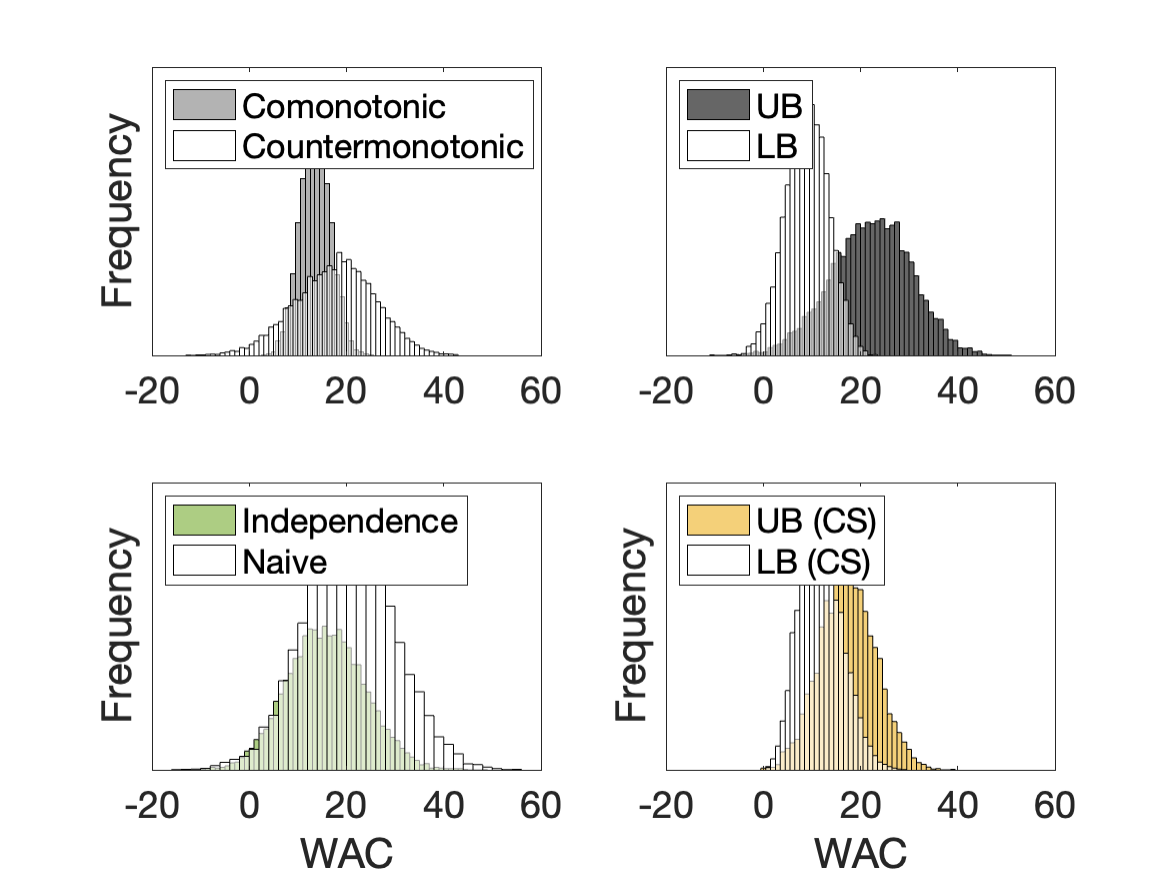}}
  \end{center}
\caption{The distribution of WAC for $\eta = 0.5$.}
\label{fig:ewac_dist}
\end{figure}

In Figures \ref{fig:ewac_dist1} and \ref{fig:ewac_dist2}, we display histograms of the WAC (obtained from the aforementioned $S=10^4$ samples) for four pairs of the SCM mechanism $\btheta$, for the value of $\eta = 0.5$ and for the same two paths that we considered in Section \ref{sec:CasinoExp}. To map Figures \ref{fig:ewac_dist1} and \ref{fig:ewac_dist2} to Figures \ref{fig:ewac1} and \ref{fig:ewac2}, note the average corresponding to each histogram in Figures \ref{fig:ewac_dist1} and \ref{fig:ewac_dist2} should match the value reported in Figures \ref{fig:ewac1} and \ref{fig:ewac2}, respectively, for $\eta = 0.5$. For example,  in Figure \ref{fig:ewac_dist1}, the comonotonic histogram has an average of approx.\ 12, which matches the reported value for the comonotonic copula when $\eta = 0.5$ in Figure \ref{fig:ewac1a}.
For each of Paths 1 and 2, the comonotonic histogram lies entirely to the right of 0, and indeed it is easy to see that $\WACP \ge 0$ w.p.\ 1 for all possible observed paths.
We observe that the comonotonic and countermonotonic histograms are very similar to the UB and LB histograms, suggesting they might be able to serve as approximations to the bounds in other applications when the bounds are difficult to compute. (Of course, one would need to provide some application-specific justification for making such an approximation.)

\end{appendices}

\end{document}